\documentclass{article}

\usepackage{microtype}
\usepackage{graphicx}
\usepackage{subfigure}
\usepackage{booktabs} 

\usepackage{hyperref}

\usepackage[accepted]{icml2023}

\usepackage{amsmath}
\usepackage{amssymb}
\usepackage{mathtools}
\usepackage{amsthm}
\usepackage{bm}

\usepackage{multirow} 

\usepackage{colortbl}
\definecolor{mygray}{gray}{.9}
\usepackage{bbm}

\usepackage[capitalize,noabbrev]{cleveref}

\theoremstyle{plain}
\newtheorem{theorem}{Theorem}[section]
\newtheorem{proposition}[theorem]{Proposition}

\theoremstyle{definition}

\theoremstyle{remark}

\usepackage[textsize=tiny]{todonotes}

\def \d1{\mathds{1}}

\def\X{\mathcal{X}}

\def\Y{\mathcal{Y}}

\def\L{\mathcal{L}}

\def \d1{\mathds{1}}

\def\1{\mathbf{1}}
\def\P{\mathbb{P}}
\def\R{\mathbb{R}}

\newcommand{\norm}[1]{\left\lVert#1\right\rVert}

\DeclareMathOperator*{\argmax}{arg\,max}
\DeclareMathOperator*{\argmin}{arg\,min}

\def\argmin{\text{argmin}}

\def\*#1{\mathbf{#1}}

\newcommand{\ignore}[1]{}

\newcommand{\abs}[1]{\left| #1 \right|}

\def\Pin{\mathbb{P}_{\text{in}}}
\def\Pwild{\mathbb{P}_{\text{wild}}}

\icmltitlerunning{Feed Two Birds with One Scone: Exploiting Wild Data for Both Out-of-Distribution Generalization and Detection}

\begin{document}

\twocolumn[
\icmltitle{Feed Two Birds with One Scone: \\
Exploiting Wild Data for Both Out-of-Distribution Generalization and Detection}



\icmlsetsymbol{equal}{*}

\begin{icmlauthorlist}
\icmlauthor{Haoyue Bai}{comp}
\icmlauthor{Gregory Canal}{ifds,ece}
\icmlauthor{Xuefeng Du}{comp}
\icmlauthor{Jeongyeol Kwon}{ece}
\icmlauthor{Robert Nowak}{comp,ece}
\icmlauthor{Yixuan Li}{comp}
\end{icmlauthorlist}


\icmlaffiliation{ifds}{Institute for Foundations of Data Science, University of Wisconsin, Madison}
\icmlaffiliation{comp}{Department of Computer Sciences, University of Wisconsin, Madison}
\icmlaffiliation{ece}{Department of Electrical and Computer Engineering, University of Wisconsin, Madison}


\icmlkeywords{Machine Learning, ICML}

\vskip 0.3in
]



\printAffiliationsAndNotice{}  

\begin{abstract}
Modern machine learning models deployed in the wild can encounter both covariate and semantic shifts, giving rise to the problems of out-of-distribution (OOD) generalization and OOD detection respectively. While both problems have received significant research attention lately, they have been pursued independently. This may not be surprising, since the two tasks have seemingly conflicting goals. This paper provides a new unified approach that is capable of simultaneously generalizing to covariate shifts while robustly detecting semantic shifts. We propose a margin-based learning framework that exploits freely available unlabeled data in the wild that captures the environmental test-time OOD distributions under both covariate and semantic shifts. We show both empirically and theoretically that the proposed margin constraint is the key to achieving both OOD generalization and detection.  Extensive experiments show the superiority of our framework, outperforming competitive baselines that specialize in either OOD generalization or OOD detection. {Code is publicly available at \url{https://github.com/deeplearning-wisc/scone}.}

\end{abstract}


\section{Introduction} \label{sec:intro}

\begin{figure*}[ht]
\centering
\includegraphics[width=0.90\textwidth]{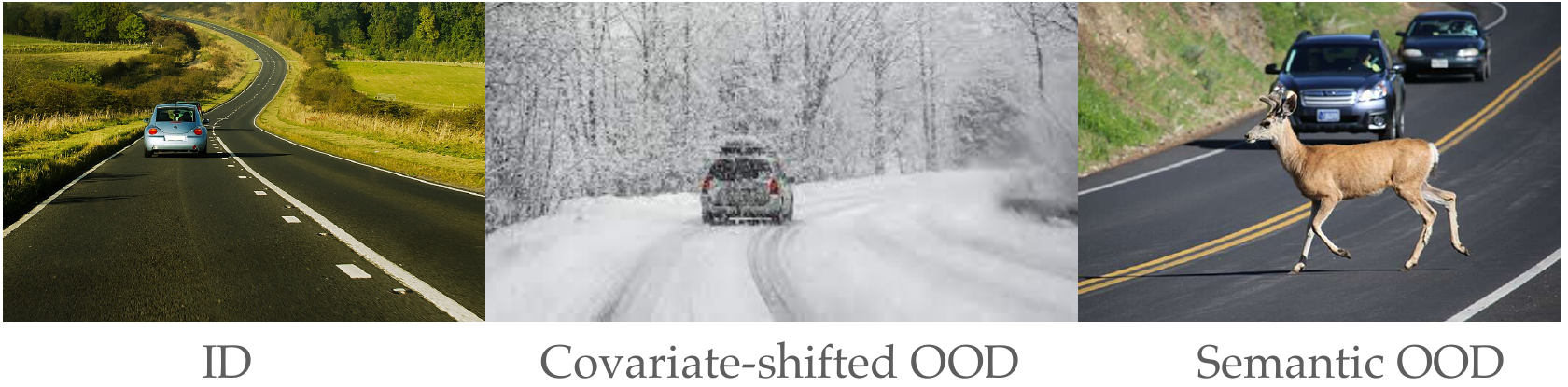}
\vspace{-0.3cm}
\caption{Illustration of three types of data that can organically arise when deploying models in the open world: (1) in-distribution (ID) data (e.g., car on a sunny day), (2) covariate-shifted OOD data (e.g., car in the snow), and (3) semantic-shifted OOD data (e.g., a deer). Our framework enables leveraging the wild mixture (containing all three types of data) for OOD generalization and OOD detection.}
\vspace{-0.3cm}
\label{fig:teaser}
\end{figure*}

Modern machine learning models deployed in the wild can encounter different types of distributional shifts. Taking autonomous driving as an example, a model trained on in-distribution (ID) data with {sunny} weather (Figure~\ref{fig:teaser}, left) may experience a \emph{covariate shift} due to {snowy} weather (Figure~\ref{fig:teaser}, middle). Under such a covariate shift, a model is expected to generalize to the out-of-distribution (OOD) data---correctly predicting  the sample into one of the known classes (e.g., car), despite the shift. Additionally, the model may encounter a \emph{semantic shift}, where samples are from unknown classes (e.g., deer) that the model has not been exposed to during training (Figure~\ref{fig:teaser}, right). Such semantic OOD data should be rejected instead of being blindly predicted as a known class.

These distributional shift scenarios give rise to the importance of two problems: OOD generalization, which focuses on the covariate shift problem~\cite{gulrajani2020search, koh2021wilds, ye2022ood}, and OOD detection, which targets semantic shift~\cite{hendrycks2016baseline,liu2020energy, yang2021generalized}. Both problems have received increasing research attention lately, albeit have been pursued independently; as a result, existing methods are highly specialized in one task, but not capable of handling both simultaneously. This has largely impeded the wider adoption
of OOD algorithms in real-world environments, which often present heterogeneous data shifts. A critical yet underexplored question thus arises:
\begin{center}
    \textbf{\emph{Can we devise a unified learning framework for both OOD generalization and OOD detection?}}
\end{center}

In this paper, we bridge the gap between OOD generalization and OOD detection, in one coherent framework. Our driving idea is to exploit unlabeled wild data naturally arising in the model's operating environment, turning the OOD threat into valuable learning resources instead. 
Wild data arises naturally \emph{for free} upon deploying any machine learning classifier in its respective environment, and has been largely overlooked for OOD learning purposes. 
Specifically, we consider a generalized characterization of the wild data, which can be modeled as a mixed composition of three data distributions:
\begin{equation*}
\mathbb{P}_\text{wild} := (1-\pi_s-\pi_c) \mathbb{P}_\text{in} + \pi_c \mathbb{P}_\text{out}^\text{covariate} +\pi_s \mathbb{P}_\text{out}^\text{semantic},
\label{eq:wild_w_cor_intro}
\end{equation*}
where $\mathbb{P}_\text{in}$, $\mathbb{P}_\text{out}^\text{covariate}$ and $\mathbb{P}_\text{out}^\text{semantic}$ denote the marginal distributions of ID, covariate-shifted OOD and semantic-shifted OOD data respectively.
Such wild data is available in abundance, does not require any human annotation, and importantly, contains the \emph{true} test time OOD distributions under both covariate and semantic shifts. Thus, our distributional model above offers {strong generality and practicality}, compared to previous works that primarily consider the semantic shift in the wild data~\cite{katz2022training}. Despite the promise, learning from such heterogeneous data is technically challenging due to the lack of clear membership (ID, Covariate-OOD, Semantic-OOD) for samples drawn from the wild data distribution $\Pwild$.

To tackle this challenge, we formulate a new learning framework \textsc{Scone}---\textbf{S}emantic and \textbf{C}ovariate \textbf{O}ut-of-distribution Lear\textbf{N}ing via \textbf{E}nergy Margins. \textsc{Scone} jointly learns a robust multi-class classifier that generalizes to covariate-OOD data, and a reliable OOD detector that detects semantic-OOD data. Our key idea is to explicitly optimize for a binary classifier based on the energy function, classifying as many samples as possible from $\P_\text{wild}$ as \textsc{out} (with positive energy), subject to two constraints (i) the ID data has energy smaller than a  negative margin value, and (ii) the multi-class classification model must maintain high accuracy. We show both theoretically (Section~\ref{sec:theory}) and empirically~(Section~\ref{sec:expe}) that the margin constraint is the key to the success of our algorithm. Intuitively, enforcing an energy margin on ID data has the effect of also lowering the energy of nearby covariate-OOD points, which are semantically related to ID points. Since lower energy increases the value of the classifier logits, the covariate-OOD points then enjoy an increased logit in their correct classes, leading to stronger OOD generalization.

Extensive experiments confirm that \textsc{Scone} can effectively improve both OOD generalization and detection performance. Compared to the most related baseline WOODS~\cite{katz2022training}, our method can substantially improve the OOD classification accuracy from $52.76\%$ to ${84.69}\%$ on covariate shifted CIFAR-10 data---a direct $\textbf{31.93}\%$ improvement. Our key contributions are:

\begin{itemize}
\vspace{-0.3cm}
\item 
To the best of our knowledge, we are among the first works that utilize wild data to jointly tackle two tasks of OOD generalization and OOD detection in one framework. 
Our problem formulation offers strong generality and practicality for real-world applications. 
\vspace{-0.2cm}
\item We propose a margin-based learning framework that exploits 
freely available, unlabeled data in the wild to solve our problem. We model wild data as a comprehensive mixture of ID samples, covariate OOD, and semantic OOD data. %
\vspace{-0.2cm}
\item We perform extensive experiments and ablations, which demonstrate the efficacy of our method. We show that \textsc{Scone} demonstrates overall strong performance in both OOD generalization and detection, outperforming baselines that specialize in one or the other.
\end{itemize}


\section{Problem Setup}\label{sec:problem_setup}

\paragraph{Labeled in-distribution data.} 
Let $\X=\R^d$ denote the input space and $\Y=\{1,\ldots, K\}$ denote the label space. We assume access to a labeled training set $\mathcal{D}_{\text{in}}^{\text{train}} = \{(\*x_i, y_i)\}_{i=1}^n$, drawn \emph{i.i.d.}\ from the joint data distribution $\P_{\mathcal{X}\mathcal{Y}}$. Let $\P_{\text{in}}$ denote the marginal distribution on $\mathcal{X}$, which is also referred to as the \emph{in-distribution}. Let $f_\theta: \X \mapsto \mathbb{R}^{K}$  denote a function for the classification task, which predicts the label of an input sample {$\*x$ as $\widehat{y}(f_\theta(\*x)) \coloneqq \argmax_{y} f_{\theta}^{(y)}(\*x)$, where $f_\theta^{(y)}(\*x)$ denotes the $y$-th element of $f_\theta(\*x)$, corresponding to label $y$.}

\begin{table*}[t]
\centering
\scalebox{0.8}{\begin{tabular}{lcccc}
\toprule
\textbf{Method}  & \textbf{OOD Accuracy (MNIST-C)} & \textbf{ID Accuracy (MNIST)} & \textbf{FPR95} & \textbf{AUROC} \\
& (OOD generalization) & (ID generalization)& (OOD detection) & (OOD detection)  \\
& $\uparrow$ & $\uparrow$ & $\downarrow$ & $\uparrow$ \\
\midrule
WOODS~\cite{katz2022training} & 88.10\% & 97.88\%  & 0.194\% & 99.88\% \\
Ours & \textbf{96.51}\%  & 97.79\% & 0.017\% & 99.99\% \\
\bottomrule
\end{tabular}}         
\vspace{-0.3cm}
\caption[]{\small WOODS~\cite{katz2022training} displays limiting OOD generalization performance. For experiments, we use 25,000 samples from the MNIST dataset as ID, and a wild mixture dataset consisting of FashionMNIST as semantic-OOD data ($\pi_s$ = 0.4) and MNIST-C~\cite{mu2019mnist} --- a covariate shifted version of MNIST --- as covariate-OOD data ($\pi_c=0.3$). 
}
\label{tab:mnist}
\end{table*}

\vspace{-0.3cm}
\paragraph{Unlabeled wild data.} Trained on the ID data, the classifier $f_\theta$ deployed into the wild can encounter various distributional shifts (see Figure~\ref{fig:teaser}). To model the realistic environment, we consider the following generalized characterization of the wild data:
\begin{equation}
\mathbb{P}_\text{wild} := (1-\pi_c-\pi_s) \mathbb{P}_\text{in} + \pi_c \mathbb{P}_\text{out}^\text{covariate} +\pi_s \mathbb{P}_\text{out}^\text{semantic} , 
\label{eq:wild_w_cor}
\end{equation}
where $\pi_c, \pi_s, \pi_c+\pi_s \in [0,1]$. Our mathematical formulation thus fully encapsulates all three possible distributions that the deployed model may encounter in practice:
\begin{itemize}
\vspace{-0.25cm}
    \item \textbf{In-distribution} $\Pin$ is the marginal distribution of  the labeled data. 
\vspace{-0.25cm}
    \item \textbf{Covariate OOD} $\mathbb{P}_\text{out}^\text{covariate}$ is the marginal distribution of $\P_{\mathcal{X'}\mathcal{Y}}$ on $\mathcal{X'}$, where the joint distribution has the same label space as the training data, yet the input space undergoes shifting in style and domain. This is relevant for \emph{OOD generalization}. 
    \vspace{-0.25cm}
      \item \textbf{Semantic OOD} $\mathbb{P}_\text{out}^\text{semantic}$: wild data that does not belong to any known categories $\mathcal{Y}=\{1,2,...,K\}$, and therefore should not be predicted by the model. This is relevant for \emph{OOD detection}. 
\end{itemize}

\vspace{-0.3cm}
\textbf{Learning goal.} Our learning framework revolves around building an OOD detector $g_\theta \colon \X \to \{\textsc{in}, \textsc{out} \}$ and multi-class classifier $f_\theta$ by leveraging data from both $\P_\text{in}$ and $\P_\text{wild}$. The OOD detector $g_\theta$ should predict semantic OOD data as \textsc{out} and otherwise predict as \textsc{in}\footnote{We use \textsc{out} to avoid abusing the term OOD. In the context of OOD detection, \textsc{out} refers particularly to ``outside the semantic space $\mathcal{Y}$''. Hence covariate-OOD falls into the \textsc{in} category, in the semantic sense.}. We notate $g_\theta$ and $f_\theta$ as sharing parameters $\theta$ to indicate the fact that these functions may share neural network parameters. In evaluating our model, we are interested in the following measurements:
\begin{align*}
& (1)~\uparrow \text{ID-Acc}(f_\theta):=\mathbb{E}_{(\*x, y) \sim \P_{\mathcal{X}\mathcal{Y}}}(\mathbbm{1}{\{{\widehat{y}(f_\theta(\*x))}=y\}}),\\
&(2)~\uparrow \text{OOD-Acc}(f_\theta):=\mathbb{E}_{(\*x, y) \sim \P_\text{out}^\text{covariate}}(\mathbbm{1}{\{{\widehat{y}(f_\theta(\*x))}=y\}}),\\
 &(3)~\downarrow \text{FPR}(g_\theta):=\mathbb{E}_{\*x \sim \P_\text{out}^\text{semantic}}(\mathbbm{1}{\{g_\theta(\*x)=\textsc{in}\}}),
	\end{align*}
where $\mathbbm{1}\{\cdot\}$ is the indicator function and the arrows indicate higher/lower is better. ID-Acc, OOD-Acc, and FPR jointly capture the \textbf{(1)} ID generalization, \textbf{(2)} OOD generalization, and \textbf{(3)} OOD detection performance, respectively. In the context of OOD detection, ID samples are considered positive, and FPR means false positive rate. 


\section{Methodology} \label{sec:method}

In this section, we present a unified learning framework that enables performing both OOD generalization and OOD detection, by way of exploiting unlabeled data in the wild.
Our framework offers substantial advantages over the counterpart approaches that rely only on the ID data, and naturally suits many applications where machine learning models are deployed in the open world.  
We start with preliminaries to lay the necessary context (Section~\ref{sec:prelim}), followed by our proposed method (Section~\ref{sec:woods_plus}) and theory~(Section~\ref{sec:theory}).

\subsection{Preliminaries}
\label{sec:prelim}

\citet{katz2022training} proposed WOODS to tackle the OOD detection problem via unlabeled wild data, which consists of the ID and semantically shifted OOD data $\mathbb{P}_\text{wild}:= (1-\pi) \mathbb{P}_\text{in} + \pi \mathbb{P}_\text{out}^\text{semantic}$. The crucial difference between our work and WOODS is whether the wild mixture data contains covariate-shifted data, which introduces new challenges not considered in prior work. As we show in this work, our formulation uniquely enables both OOD generalization and OOD detection, in one coherent framework.

As preliminaries,
WOODS minimizes the error of declaring data from $\P_\text{wild}$ as ID, subject to \emph{(i)} the error of declaring an ID point as OOD is at most a fixed threshold $\alpha$, and \emph{(ii)} the multi-class classification model meets some error threshold $\tau$. Mathematically, this can be formalized as a constrained optimization problem:
    \begin{align}
        \argmin_\theta & ~ \mathbb{E}_{\*x \sim \Pwild}(\mathbbm{1}{\{g_\theta(\*x)=\textsc{in}\}}) \label{eq:infinite_sample_objective} \\
        \text{s.t. } &  \mathbb{E}_{\*x \sim \Pin}(\mathbbm{1}{\{g_\theta(\*x)=\textsc{out}\}}) \leq \alpha \nonumber \\
        & \mathbb{E}_{(\*x, y) \sim \P_{\mathcal{X}\mathcal{Y}}}(\mathbbm{1}{\{{\widehat{y}(f_\theta(\*x))}\ne y\}}) \leq \tau. \nonumber
    \end{align}
In particular, the OOD detector $g_\theta$ is defined based on the level set: $g_\theta(\*x) = \textsc{out}$ if ${E_\theta(\*x) > 0}$, where the free energy $E_\theta(\*x) := -\log \sum_{j=1}^K e^{f_{\theta}^{(j)}(\*x)}$ was shown to be an effective OOD score~\cite{liu2020energy}. ID data tends to have negative energy and vice versa.

In practice, the objective in \eqref{eq:infinite_sample_objective} can be empirically optimized over \emph{i.i.d.}\ samples $\widetilde{\*x}_1 \dots \widetilde{\*x}_m \sim \Pwild$ and $\*x_1 \dots \*x_n \sim \Pin$ via a tractable relaxation by replacing the $0/1$ loss with a surrogate loss as follows: 
\begin{align}
       \argmin_\theta \: & \frac{1}{m} \sum_{i=1}^{m} \L_\text{ood} (g_\theta(\tilde{\*x}_i), \textsc{in}) \label{eq:classification_approach} \\ 
    \text{s.t. } & \frac{1}{n} \sum_{j=1}^n \L_\text{ood} (g_\theta({\*x}_j), \textsc{out}) \leq \alpha \nonumber \\
    &  \frac{1}{n} \sum_{j=1}^n \L_\text{cls}(f_\theta(\*x_j), y_j) \leq \tau, \nonumber
\end{align}
where $\L_\text{ood}(g_\theta(\*x_i), \textsc{out}) = \frac{1}{1+\exp(-w \cdot E_{\theta} (\*x_i))}$ denotes the loss of the binary OOD classifier {(where $w \in \R$ is a learnable parameter)} and 
$\mathcal{L}_\text{cls} (f_\theta(\*x),y)$ is the per-sample cross-entropy (CE) loss for the classification task.

\textbf{Limitation in OOD generalization performance {of WOODS: a} case study.} {Although WOODS can simultaneously learn an  OOD detector and an ID classifier, it can perform poorly on the task of OOD generalization.} To see this, we investigate the efficacy of using the constrained optimization in Equation~\eqref{eq:classification_approach} directly for our new problem setting described in Section~\ref{sec:problem_setup}. To simulate the wild data $\Pwild$, we mix a subset of MNIST training data (as $\P_\text{in}$) with MNIST-C ~\cite{mu2019mnist} data --- a covariate-shifted version of MNIST --- as $\P_\text{out}^\text{covariate}$, and the FashionMNIST dataset~\cite{xiao2017fashion} as $\P_\text{out}^\text{semantic}$. We train a version of WOODS on this data and summarize our findings in Table~\ref{tab:mnist}: interestingly, we observe a significant generalization gap between classifying ID data (97.88\% in accuracy) and covariate-shifted OOD data (88.10\% in accuracy). This suggests that the training objective in~\cite{katz2022training} is indeed insufficient for the purpose of OOD generalization, despite strong OOD detection performance. This motivates our method which accounts for the task of OOD generalization on covariate-shifted data.

\begin{figure}[t]
\centering
\subfigure[No margin ($\eta = 0$)]{\includegraphics[width=0.235\textwidth]{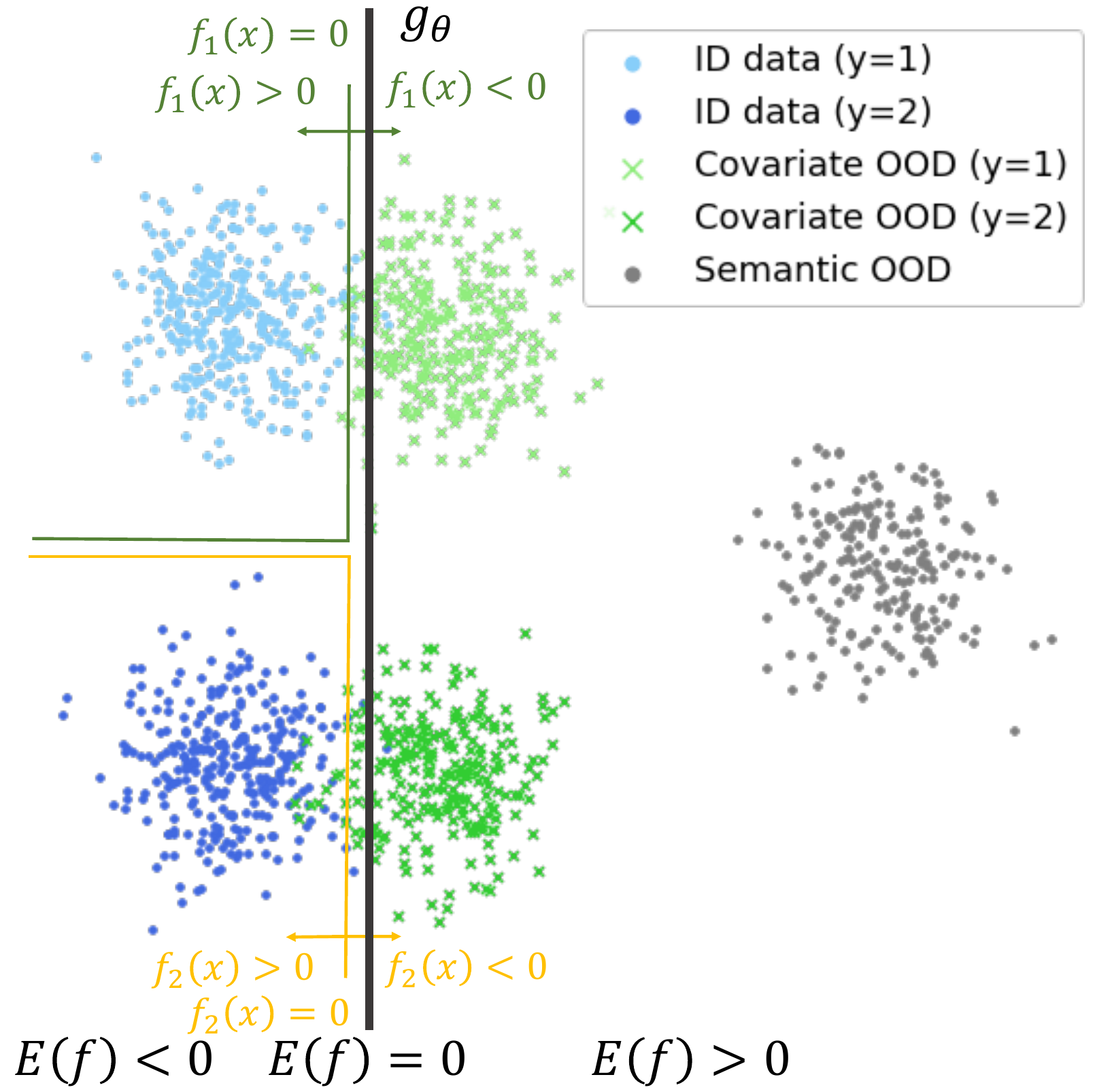}}%
\hspace{1mm}
\subfigure[Energy margin ($\eta < 0$)]{\includegraphics[width=0.235\textwidth]{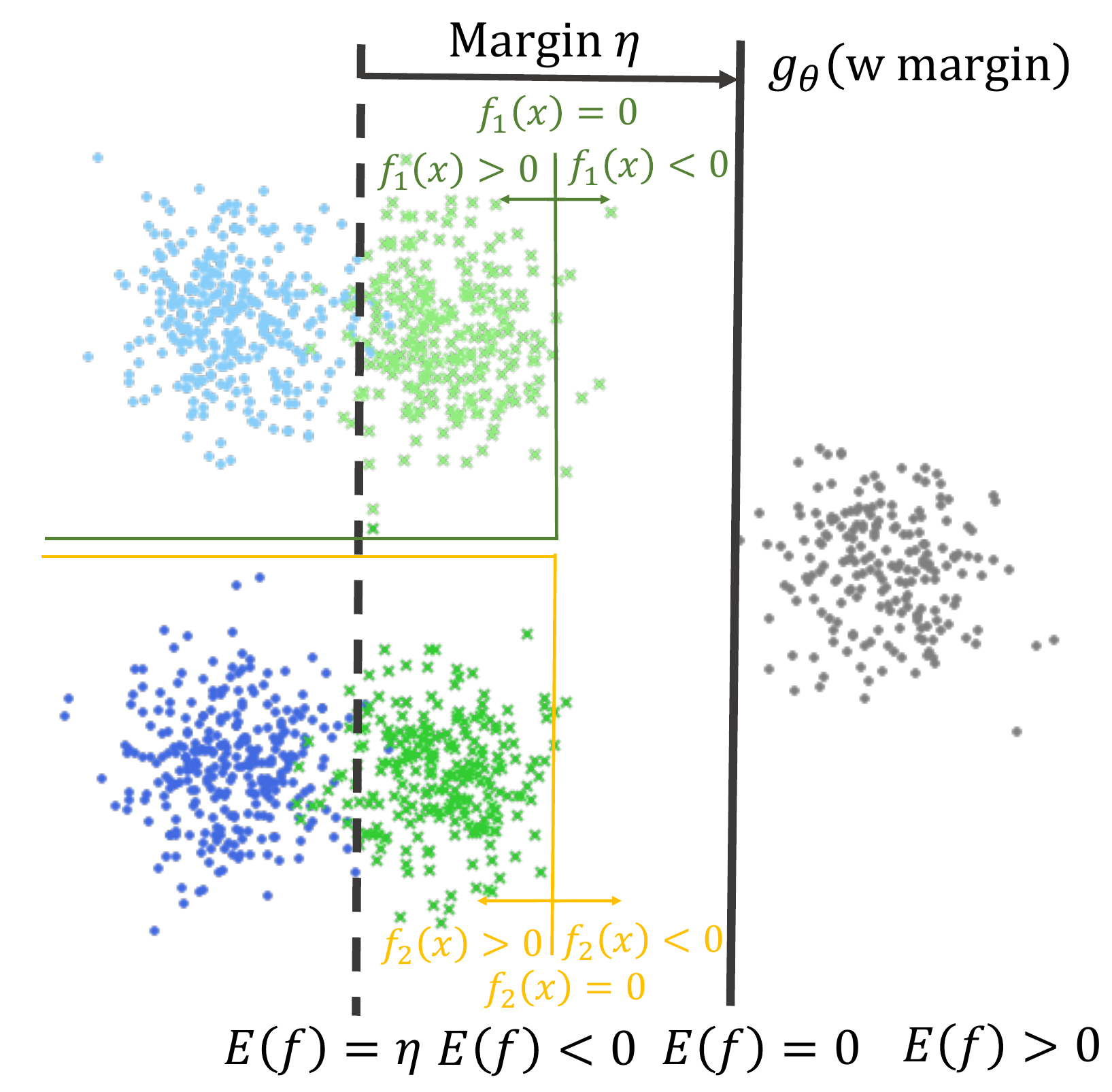}}%
\caption{Illustration of the impact of energy margin $\eta$ on the placement of the OOD detection boundary. 
}
\vspace{-0.5cm}
\label{fig:illustrate}
\end{figure}

\subsection{Proposed Method}
\label{sec:woods_plus}
\paragraph{Motivation.} Before we present our method, an important step is to identify the key reason for the limited OOD generalization performance observed when using WOODS. To better understand the behavior of WOODS, we use a simple {two-class} example {($K = 2$)} to illustrate the decision boundary of its {energy-based} OOD detector $g_\theta(\*x) = \textsc{out}$ if ${E_\theta(\*x) > 0}$ (black
vertical line in Figure~\ref{fig:illustrate}a). Geometrically, since $E_\theta(\*x)$ is high when $f_\theta^{(1)}$ and $f_\theta^{(2)}$ are both low, the objective in \eqref{eq:infinite_sample_objective} attempts to minimize $f_\theta(\*x)$ elementwise on points in $\Pwild$. Conversely, since $f_\theta(\*x)$ is simultaneously being trained to classify ID points (in \textcolor{blue}{blue}) correctly, it will be optimized towards being ``one-hot'' on each point in $\Pin$.

With this perspective, our critical insight is that the objective in Equation~\eqref{eq:infinite_sample_objective} incentivizes the OOD detector to be as close as possible to the ID data (colored in \textcolor{blue}{blue}), since it aims to classify as many samples as possible from $\Pwild$ {--- which includes covariate-shifted data in this setting ---} as semantic \textsc{out}. Therefore, the OOD detection boundary undesirably places the covariate-shifted data (colored in
\textcolor{green}{green})  on the wrong side --- labeling it as semantic \textsc{out} --- {and drives its label distribution towards uniform (since neither $f_\theta^{(1)}$ or $f_\theta^{(2)}$ are ``one-hot''), resulting in classification errors.} Instead, the OOD detector should ideally place only the semantically shifted OOD data (colored in \textcolor{gray}{gray}) on the \textsc{out} side of the detection boundary.

\textbf{{Proposed margin-based learning objective.}} Leveraging these insights, we now propose a new learning objective to mitigate this issue. \textsc{Scone} is motivated by the observation in Figure~\ref{fig:illustrate}a, where the OOD decision boundary lacks the sufficient margin \emph{w.r.t.}\ the ID data. Therefore, our key idea is to enforce a sufficient margin between the OOD detector and the ID data. %
Our key idea is to enforce the \emph{ID data to have energy smaller than the margin $\eta$} (a negative value), while optimizing for the level-set estimation based on the energy function. Given access to samples $\tilde{\*x}_{1},\ldots, \tilde{\*x}_{m}$ from $\Pwild$, along with labeled ID samples $(\*x_1, y_1), \ldots, (\*x_n,y_n)$, our proposed optimization is:
\begin{align}
            \argmin_{\theta} \: & \frac{1}{m}\sum_{i=1}^{m} \mathbbm{1}\{ E_\theta(\tilde{\*x}_i) \leq 0\}  \label{eq:energy_robust01}\\
         \text{s.t. } & \frac{1}{n}\sum_{j=1}^n \mathbbm{1}\{ E_\theta(\*x_j) \geq \eta\} \leq \alpha \notag\\
         &  \frac{1}{n} \sum_{j=1}^n \mathbbm{1}\{{\widehat{y}(f_\theta(\*x_j))}\ne y_j\} \leq \tau, \notag
\end{align}
where $\eta$ controls the margin of the OOD decision boundary \emph{w.r.t.}\ the ID data. Note that our learning objective generalizes WOODS~\cite{katz2022training}, which corresponds to the special case of $\eta=0$ (i.e., no margin at all). 

{We illustrate the intuitive effect of this ID energy margin constraint in Figure~\ref{fig:illustrate}b: by requiring lower energy on ID points with a hard constraint in \eqref{eq:energy_robust01}, the OOD detector $g_\theta$ is forced to move its zero level-set to the right in order to decrease ID energy. Since decreasing $E_\theta(\*x)$ directly corresponds to increasing $f_\theta^{(1)}(\*x)$ and/or $f_\theta^{(2)}(\*x)$, this is achieved by moving the zero level-sets of $f_\theta^{(1)}(\*x)$ and $f_\theta^{(2)}(\*x)$ to the right. Since $f_\theta^{(1)}(\*x)$ and $f_\theta^{(2)}(\*x)$ are also constrained to classify ID data correctly, we expect that this level-set shift will still preserve the classifier boundary between ID points, resulting in the geometry visualized in Figure~\ref{fig:illustrate}b. Crucially, due to the assumed nearness of covariate-shifted points to ID points, these shifts can have the effect of then generalizing the ID classification boundary to $\mathbb{P}_\text{out}^\text{covariate}$. Although \eqref{eq:energy_robust01} is also attempting to maximize the number of points from $\mathbb{P}_\text{out}^\text{covariate}$ detected as \textsc{out} (i.e., on the right side of the $E_\theta(\*x) = 0$ boundary), this is a weaker force than the ID energy margin, since the latter is a hard constraint. To provide some theoretical insights, we next formalize these intuitions for a restricted model architecture.}

\subsection{{Theoretical Insights: Relationship Between OOD Generalization and Energy Margin}}
\label{sec:theory}
We now discuss how \textsc{Scone} can \emph{improve the classification} accuracy on the covariate-shifted data under some assumptions on the data distribution and model class. 
Suppose we are given a fixed feature map $\phi \colon \R^d \to \R^p$ to some feature space in $\R^p$, where there exists a constant $\delta > 0$ such that for each covariate-shifted point $\*x_c$ with ground-truth label $y$, there exists a corresponding ID point $\*x$, also with label $y$, satisfying $\norm{\phi(\*x_c) - \phi(\*x)}_2 < \delta$. That is, we assume that the covariate-shifted data is close to in-distribution data in the feature space, which is supported empirically in our experiments (Section~\ref{expe:visualize}). 
Suppose a classifier $f_\theta \colon \R^p \to \R^K$ is learned on top of $\phi(\*x)$, using the ``idealized'' version of our method in \eqref{eq:energy_robust01} (replacing each data point $\*x_i, \widetilde{\*x}_i$ in \eqref{eq:energy_robust01} with its corresponding feature vector $\phi(\*x_i), \phi(\widetilde{\*x}_i)$), and consider the two-class case ($K=2$). Since each classification decision only depends on the difference $f_\theta^{(1)}(\cdot) - f_\theta^{(2)}(\cdot)$, for analytical convenience suppose that we learn this difference directly as $\overline{f}_\theta(\cdot) = f_\theta^{(1)}(\cdot) - f_\theta^{(2)}(\cdot)$ with $\overline{f}_\theta(\cdot) > 0$ corresponding to $y=1$ and $\overline{f}_\theta(\cdot) < 0$ indicating $y=2$, which we can accomplish in our framework by fixing $f_\theta^{(2)} = -\frac12 \overline{f}_\theta$ and $f_\theta^{(1)} = \frac12 \overline{f}_\theta$. 

Suppose further that we set $\alpha = 0$ and $\tau = 0$, such that every ID point is classified correctly and has energy satisfying $E_\theta(\phi(\*x)) < \eta$. Finally, suppose that $\overline{f}_\theta$ is $L$-Lipschitz: this is the case for many classifier functions, such as two-layer ReLU networks with bounded variation \cite{parhi2022what}. {We then have the following result on the covariate-shifted points (proved in Appendix \ref{app:proof}):}%
\begin{proposition}
\label{prop:covcorrect}

(Informal) {Under some assumptions,} 
{if} $\eta < - \log2 - \frac12 L \delta$ {then} each covariate-shifted point is classified correctly {and is detected as semantic \textsc{in}}.
\end{proposition}

\vspace{-0.5cm}
\textbf{Implications.} This result illustrates that even though our method does not have access to the ground-truth labels of covariate-shifted data during training, we expect that as long as each covariate-shifted data point is ``close'' to a corresponding ID point, then setting the ID energy threshold $\eta$ appropriately will result in the covariate-shifted data being classified \emph{correctly}. Intuitively, requiring the ID data points to have lower energy while simultaneously being classified correctly encourages their logit values $f_\theta(\cdot)$ to move further from the classification decision boundary, on the correct side. If $f_\theta$ belongs to a regularized function class (e.g., Lipschitz functions), then the logits of the covariate-shifted data will not deviate wildly from their ID counterparts. Combining these two insights, the logits of the covariate-shifted data should also be bounded away from the decision boundary, on the correct side. Importantly, this distance to the boundary is explicitly increased by decreasing $\eta$, whereas WOODS does not necessarily ensure this property.

\subsection{{Enforcing Margin in Practice}}
Since the $0/1$ loss {in \eqref{eq:energy_robust01}} is intractable, {in a similar manner to WOODS}, we replace it with a smooth approximation given by the binary sigmoid loss, yielding the following optimization problem:
\begin{align}
\argmin_{\theta, w \in \R} \: & \frac{1}{m}\sum_{i=1}^{m} \frac{1}{1 + \exp(w \cdot E_\theta(\tilde{\*x}_i))} \label{eq:energy_robust} \\
        \text{s.t. } & \frac{1}{n}\sum_{j=1}^n \frac{1}{1+\exp(-w \cdot (E_{\theta} (\*x_j)-\eta))}   \leq \alpha \nonumber \\
       & \frac{1}{n} \sum_{j=1}^n \L_\text{cls}( f_{\theta}(\*x_j) , y_j) \leq \tau, \nonumber
\end{align}
\textbf{Solving the constrained optimization.} We adopt the Augmented Lagrangian method~\cite{hestenes1969multiplier} to solve our constrained optimization problem with modern neural networks. In short, the constrained optimization problem above is converted into a sequence of unconstrained optimization problems. We refer interested readers to Section 3.2 in~\citet{katz2022training} for details. 
We showcase the efficacy of our algorithm on the simple MNIST example in Table~\ref{tab:mnist}, where the proposed method improves the OOD generalization accuracy from 88.10\% (WOODS) to \textbf{96.51}\%. Building on this encouraging result, we proceed to comprehensively evaluate our algorithm in the next section.


\vspace{-0.2cm}
\section{Experiments} \label{sec:expe}

In this section, we comprehensively verify the empirical efficacy of \textsc{Scone}. We first describe the experimental setup (Section~\ref{expe:setup}). In Section~\ref{expe:results}, we present results for both OOD generalization and OOD detection, followed by extensive ablations. We provide qualitative analysis that improves the understanding of \textsc{Scone} in Section~\ref{expe:visualize}.

\subsection{Experimental Setup} \label{expe:setup}

\paragraph{Datasets and evaluation metrics.} 
Following the common benchmarks in literature, we use CIFAR-10~\cite{krizhevsky2009learning}  as the in-distribution data ($\Pin$). For the covariate-shifted data ($\mathbb{P}_\text{out}^\text{covariate}$), we use CIFAR-10-C~\cite{hendrycks2018benchmarking}
with Gaussian additive noise for our main experiments, and provide ablations in  Appendix~\ref{app:corruption} on other types of covariate shifts. For semantic-shifted OOD data ($\mathbb{P}_\text{out}^\text{semantic}$), we use natural image datasets: SVHN \cite{netzer2011reading}, Textures \cite{cimpoi2014describing}, Places365 \cite{zhou2017places}, LSUN-Crop \cite{yu2015lsun}, and LSUN-Resize \cite{yu2015lsun}. {Large-scale results on the ImageNet dataset can be found in Section~\ref{app:imagenet}. {Additional results on the PACS dataset~\cite{li2017deeper} from DomainBed is presented in Appendix~\ref{app:pacs}}. We provide a detailed description of the datasets in Appendix~\ref{app:datasets}.}

To simulate the wild data $\Pwild$, we mix a subset of ID data ($\P_\text{in}$) with the covariate shifted dataset ($\P_\text{out}^\text{covariate}$) under various $\pi_c \in \{0.0, 0.1, 0.2, 0.5, 0.9\}$; we default to $\pi_c=0.5$ for the main experiment. We keep $\pi_s=0.1$ to reflect the fact that we expect semantic OOD shifts ($\P_\text{out}^\text{semantic}$) to be encountered less frequently, which is the same value as used in~\cite{katz2022training}. We split ID datasets into two halves: we use 50\% as ID training data and 50\% for creating the mixture data.

In each training iteration, we simulate the
mixture data as follows. For the ID dataset we draw one batch of size $128$, and for the wild dataset $\P_\text{wild}$ we draw another batch of size $128$ where each example is drawn from $\mathbb{P}_\text{out}^\text{covariate}$ with probability $\pi_c$, from $\mathbb{P}_\text{out}^\text{semantic}$ with probability $\pi_s$ and from $\P_\text{in}$ with probability $1-\pi_c-\pi_s$. For evaluation, we use the original test split of the ID data. Details of data split for OOD datasets are described in Appendix~\ref{app:datasets}. 
To evaluate each method including baselines, we use the collection of metrics defined in Section~\ref{sec:problem_setup}. The threshold for an energy-based OOD detector is selected based on the ID test set when 95\% of ID test data points are declared as ID.

\textbf{Training details.} 
For CIFAR experiments and methods, we use the Wide ResNet~\cite{zagoruyko2016wide} architecture with 40 layers and widen factor of 2. The model is optimized using stochastic gradient descent with Nesterov momentum \cite{duchi2011adaptive}. 
We set the weight decay as 0.0005, and momentum as 0.09. We initialize the model with a pre-trained network on {CIFAR-10}, and then trained for 100 epochs using our method. The initial learning rate is 0.0001 and decays by a factor of 2 at epochs 50, 75, and 90. Following the previous literature~\cite{katz2022training}, we set $\alpha=0.05$, and set $\tau$ to be twice the loss of the pre-trained model. For all the experiments, we use a batch size of 128 and a dropout rate of 0.3. 
Our framework was implemented with PyTorch 1.8.1.  All training is performed using NVIDIA GeForce RTX 2080 Ti. See Appendix~\ref{app:experimental_details} for additional experimental details, including our validation strategy for selecting $\eta$.

\subsection{Results and Discussion} \label{expe:results}

\textbf{Effect of margin $\eta$.} Since the energy margin is central to our learning framework, we first aim to understand how $\eta$ impacts performance. In Table~\ref{tab:margin}, we perform an ablation by varying the margin
$\eta \in \{0, -0.1, -0.5, -1, -2, -10, -20, -50\}$. Since energy should be negative for ID data, more negative values of $\eta$ translate to a stronger margin constraint on ID points. As the margin changes from $\eta=0$ to $\eta=-10$, a salient observation is that our method can substantially improve the OOD accuracy from $52.76\%$ to ${84.69}\%$---a direct $\textbf{31.93}\%$ improvement in accuracy. At the same time, the ID accuracy remains  comparable across different $\eta$, suggesting that one can indeed leverage the wild data to gain an improvement in OOD generalization  without sacrificing ID classification accuracy. Furthermore, we observe that a tradeoff may exist between OOD generalization and OOD detection performance: the optimal OOD accuracy is achieved under a large margin, which can slightly degrade the OOD detection performance when compared to $\eta=0$. Despite the tradeoff, we will show that \textsc{Scone} still outperforms competing OOD detection methods (\emph{c.f.} Table~\ref{tab:ood}).

\begin{table}[t]
\centering
\vspace{-0.25cm}
\caption{Experimental results on CIFAR with different margin $\eta$. We train on CIFAR-10 as ID, using wild data with $\pi_c=0.5$ (CIFAR-10-C) and $\pi_s=0.1$ (SVHN).}
\scalebox{0.9}{
\begin{tabular}{c|cccc}
\toprule
\textbf{margin}  
& \textbf{OOD Acc.}$\uparrow$ & \textbf{ID Acc.}$\uparrow$ &\textbf{FPR}$\downarrow$ & \textbf{AUROC}$\uparrow$   \\
\midrule
No margin  & 52.76 & 94.86 & 2.11 & 99.52  \\
$\eta = -0.1$ & 53.24 & 94.87 & 2.16 & 99.52  \\
$\eta = -0.5$  & 54.22 & 94.85 & 2.31 & 99.49 \\
$\eta = -1$  & 55.55  & 94.88 & 2.56 & 99.45  \\
$\eta = -2$  & 58.47 & 95.00 & 3.19 & 99.35  \\
\rowcolor{mygray} $\eta = -10$ & 84.69 & 94.65 & 10.86 & 97.84  \\
$\eta = -20$ & 84.57 & 94.81 & 19.04 & 96.29  \\
$\eta = -50$ & 84.56 & 94.83 & 19.24 & 96.25  \\
\bottomrule
\end{tabular}%
}
\vspace{-0.6cm}
\label{tab:margin}%
\end{table}%

\begin{table*}[ht]
\centering
\caption{Main results: comparison with competitive OOD generalization and OOD detection methods on CIFAR-10. We run our method 3 times and report the average and std. For experiments using $\mathbb{P}_{\mathrm{wild}}$, we set $\pi_s = 0.5$, $\pi_c = 0.1$. For each semantic OOD dataset, we create corresponding wild mixture distribution $\mathbb{P}_\text{wild} := (1-\pi_s-\pi_c) \mathbb{P}_\text{in} + \pi_s \mathbb{P}_\text{out}^\text{semantic} + \pi_c \mathbb{P}_\text{out}^\text{covariate}$ for training and evaluating on the corresponding test dataset. $\pm x$ denotes the standard error, rounded to the first decimal point.
Results for LSUN-R and Texture datasets are in  Appendix~\ref{app:additional_ood}. (*Since all the OOD detection methods use the same model trained with the CE loss on $\Pin$, they display the same ID and OOD accuracy on CIFAR-10-C.)}
\scalebox{0.65}{
\begin{tabular}{lcccc|cccc|cccc}
\toprule
\multirow{2}[2]{*}{\textbf{Method}} & \multicolumn{4}{c}{{SVHN $\mathbb{P}_\text{out}^\text{semantic}$, CIFAR-10-C $\mathbb{P}_\text{out}^\text{covariate}$}} & \multicolumn{4}{c}{{LSUN-C $\mathbb{P}_\text{out}^\text{semantic}$, CIFAR-10-C $\mathbb{P}_\text{out}^\text{covariate}$}} & \multicolumn{4}{c}{{Places365 $\mathbb{P}_\text{out}^\text{semantic}$, CIFAR-10-C $\mathbb{P}_\text{out}^\text{covariate}$}}  \\
 & \textbf{OOD Acc.}$\uparrow$  & \textbf{ID Acc.}$\uparrow$ & \textbf{FPR}$\downarrow$ & \textbf{AUROC}$\uparrow$ & \textbf{OOD Acc.}$\uparrow$ & \textbf{ID Acc.}$\uparrow$ & \textbf{FPR}$\downarrow$ & \textbf{AUROC}$\uparrow$ &  \textbf{OOD Acc.}$\uparrow$ & \textbf{ID Acc.}$\uparrow$ & \textbf{FPR}$\downarrow$ & \textbf{AUROC}$\uparrow$ \\
\midrule
\emph{OOD detection}\\
\textbf{MSP}  & 75.05 & 94.84 & 48.49 & 91.89 & 75.05 & 94.84 & 30.80 & 95.65 & 75.05 & 94.84 & 57.40 & 84.49 \\
\textbf{ODIN}  & 75.05 &  94.84 & 33.35 & 91.96 & 75.05 &  94.84  & 15.52 & 97.04 & 75.05 &  94.84  & 57.40 & 84.49 \\
\textbf{Energy}  & 75.05 &  94.84  & 35.59 & 90.96 & 75.05 &  94.84  & 8.26 & 98.35 & 75.05 &  94.84  & 40.14 & 89.89 \\
\textbf{Mahalanobis}  & 75.05 &  94.84  & 12.89 & 97.62 & 75.05 &  94.84 & 39.22 & 94.15 & 75.05 &  94.84 & 68.57 & 84.61 \\
\textbf{ViM} & 75.05 & 94.84 & 21.95 & 95.48 & 75.05 & 94.84 & 5.90 & 98.82 & 75.05 & 94.84 & 21.95 & 95.48 \\
\textbf{KNN} & 75.05 & 94.84 & 28.92 & 95.71 & 75.05 & 94.84 & 28.08 & 95.33 & 75.05 & 94.84 & 42.67 & 91.07 \\
\midrule
\emph{OOD generalization}\\
\textbf{ERM } & 75.05 &  94.84  & 35.59 & 90.96 & 75.05 &  94.84  & 8.26 & 98.35 & 75.05 &  94.84  & 40.14 & 89.89 \\
\textbf{Mixup }  & 79.17 & 93.30 & 97.33 & 18.78 & 79.17 & 93.30 & 52.10 & 76.66 & 79.17 & 93.30 & 58.24 & 75.70 \\
\textbf{IRM }  & 77.92 & 90.85 & 63.65 & 90.70 & 77.92 & 90.85 & 36.67 & 94.22 & 77.92 & 90.85 & 53.79 & 88.15 \\
\textbf{VREx }  & 76.90 & 91.35 & 55.92 & 91.22 & 76.90 & 91.35 & 51.50 & 91.56 & 76.90 & 91.35 & 56.13 & 87.45 \\
\midrule
\emph{Learning w. $\Pwild$}\\
\textbf{OE}         & 37.61 & 94.68 & 0.84 & 99.80 & 41.37 & 93.99 & 3.07 & 99.26  & 35.98 & 94.75 & 27.02 & 94.57    \\
\textbf{Energy (w. outlier)} & 20.74 & 90.22 & 0.86 & 99.81 & 32.55 & 92.97 & 2.33 & 99.93  & 19.86 & 90.55 & 23.89 & 93.60    \\
\textbf{Woods} & 52.76  & 94.86 & 2.11 & 99.52 & 76.90 & 95.02 & 1.80 & 99.56  & 54.58 & 94.88  & 30.48 & 93.28   \\
\rowcolor{mygray}
\textbf{Scone} (ours) & 84.69$_{\pm 0.1}$ & 94.65$_{\pm 0.0}$  & 10.86$_{\pm 0.7}$ & 97.84$_{\pm 0.1}$  & 84.58$_{\pm 0.7}$ & 93.73$_{\pm 0.4}$  & 10.23$_{\pm 1.1}$ & 98.02$_{\pm 0.2}$ & 85.21$_{\pm 0.1}$ & 94.59$_{\pm 0.0}$  & 37.56$_{\pm 0.2}$ & 90.90$_{\pm 0.1}$  \\
\bottomrule
\end{tabular}%
}
\label{tab:ood}%
\end{table*}%

\vspace{0.2cm}
\textbf{\textsc{Scone} achieves strong performance.} 
We present the main results in Table~\ref{tab:ood}, where \textsc{Scone} establishes \emph{overall strong performance in both OOD generalization and OOD detection}.  In particular, we consider two broad categories of methods that are developed for either OOD detection or OOD generalization, and thus are expected to excel in only one of these two tasks. In contrast, our method targets both tasks simultaneously. 
Details of baseline implementation are  in Appendix~\ref{app:experimental_details}.

We highlight a few observations:
\textbf{(1)} \textsc{Scone} outperforms competitive post hoc OOD detection methods, including \texttt{MSP}~\cite{hendrycks2016baseline}, \texttt{ODIN}~\cite{liang2018enhancing}, 
\texttt{Energy}~\cite{liu2020energy}, 
\texttt{Mahalanobis}~\cite{lee2018simple}, 
~\texttt{ViM}~\cite{wang2022vim},
and the latest baseline \texttt{KNN}~\cite{sun2022out} --- all of which use the same model trained with the CE loss on $\Pin$ (and hence they display the same OOD accuracy on CIFAR-10-C).  As an example of our method's improved performance, when using SVHN as $\mathbb{P}_\text{out}^\text{semantic}$ our method yields an FPR95 of 10.86\% (lower is better), which outperforms the best baseline performance of 12.89\%. At the same time, the OOD generalization performance is significantly improved from 75.05\% (baseline CE model) to 84.69\% (ours). \textbf{(2)} Our method also outperforms common OOD generalization baselines, including \texttt{IRM}~\cite{arjovsky2019invariant}, \texttt{Mixup}~\cite{zhang2018mixup}, and \texttt{VREx}~\cite{krueger2021out}.  While these methods display stronger OOD generalization performance than the \texttt{ERM} (or \texttt{CE}) baseline, they underperform ours both in terms of OOD generalization and OOD detection.\footnote{Same as ours, we use the energy score in test time to perform OOD detection.} \textbf{(3)} Lastly, we also compare to OOD detection methods utilizing $\Pwild$, including \textsc{oe}~\cite{hendrycks2018deep}, energy-regularized learning~\cite{liu2020energy}, and WOODS~\cite{katz2022training}, which is the latest such method to be developed. These methods are among the strongest OOD detection methods, yet display a significantly worsened OOD generalization performance. The main reason is that they make assumptions on $\Pwild$ without considering covariate OOD data.

\begin{figure*}[t]
\centering
\subfigure[Energy scores of WOODS]{\includegraphics[width=0.232\textwidth]{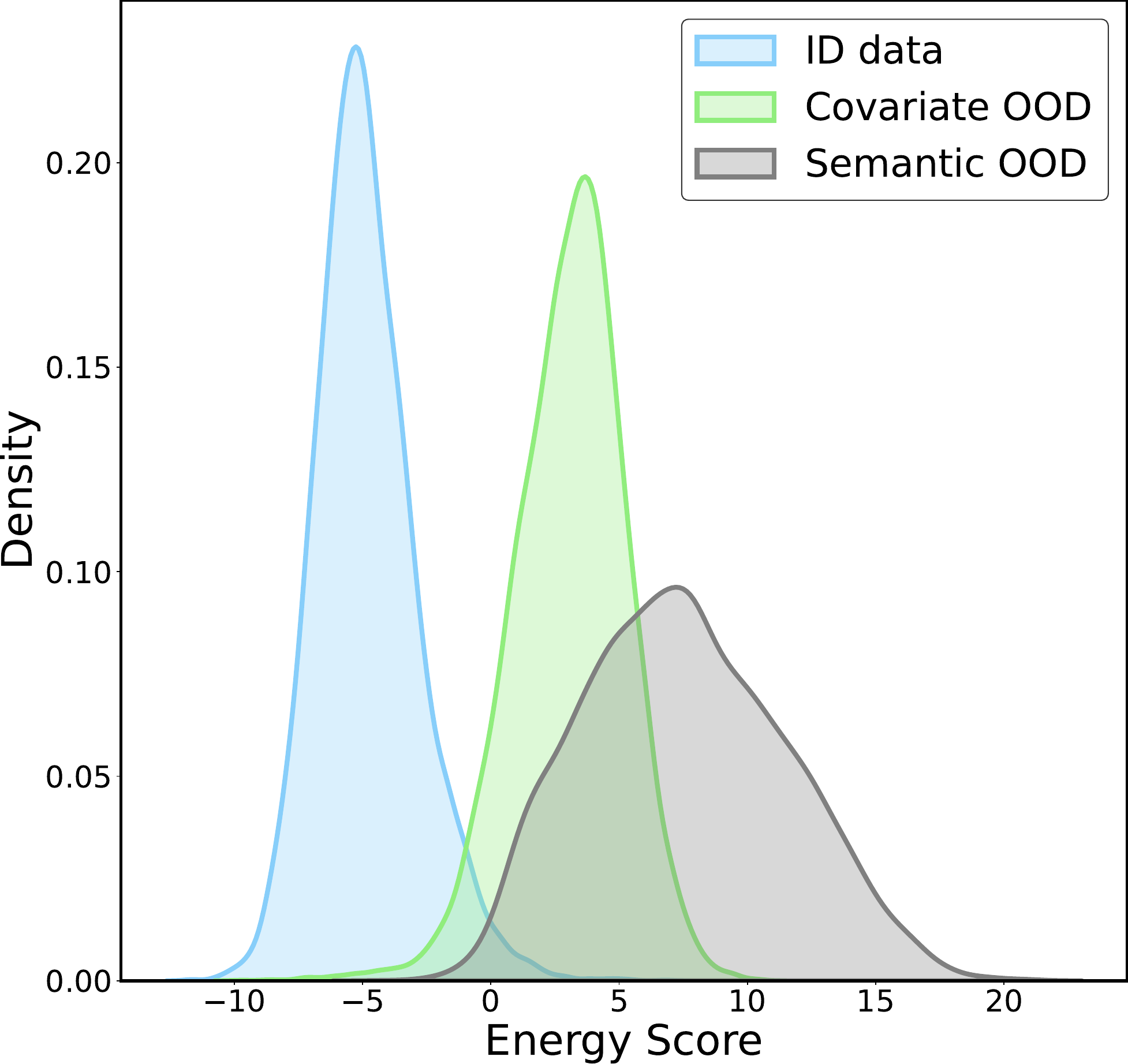}}
\subfigure[Energy scores of Ours]{\includegraphics[width=0.232\textwidth]{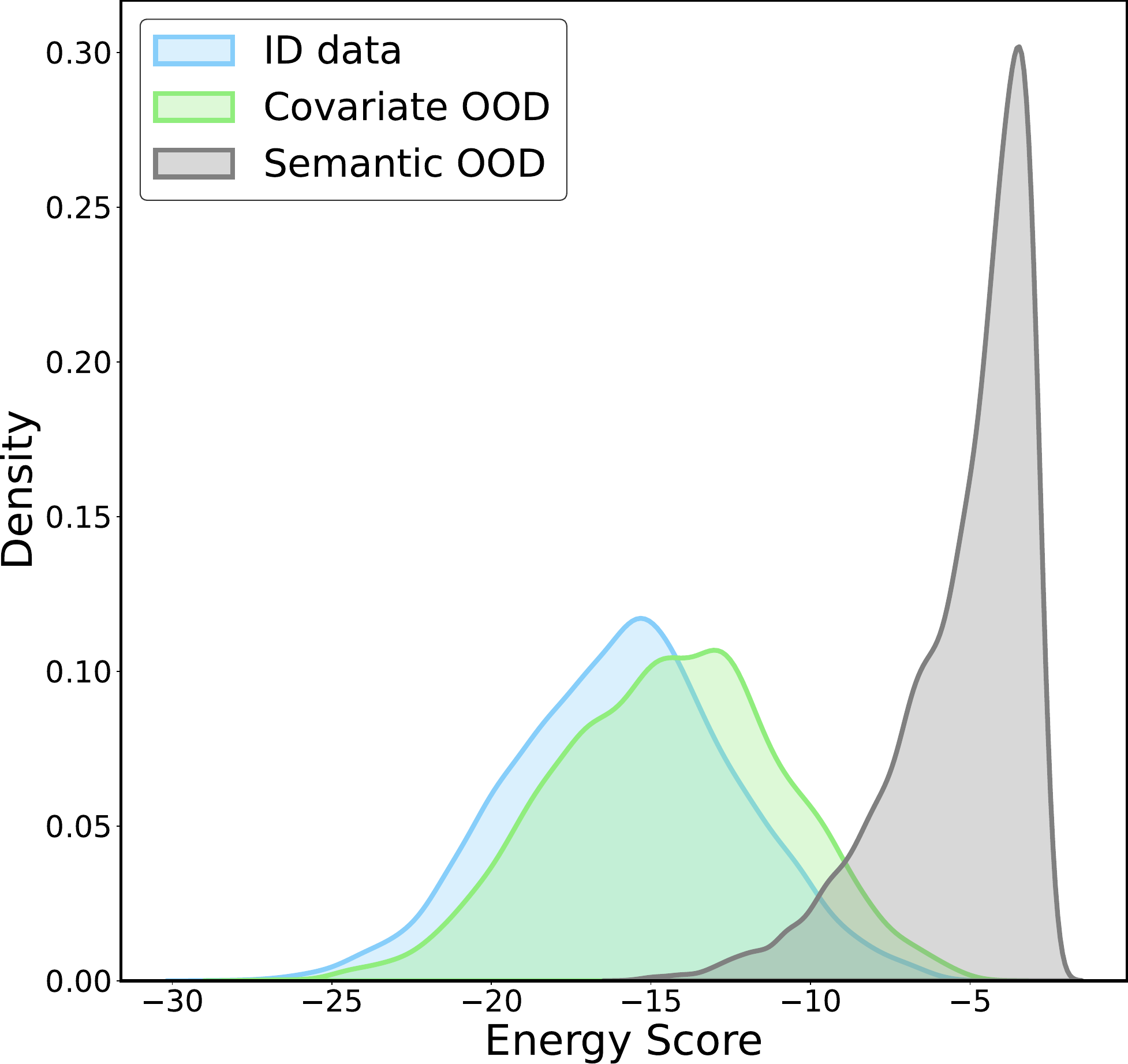}}
\hspace{1.5mm}
\subfigure[T-SNE of WOODS]{\includegraphics[width=0.22\textwidth]{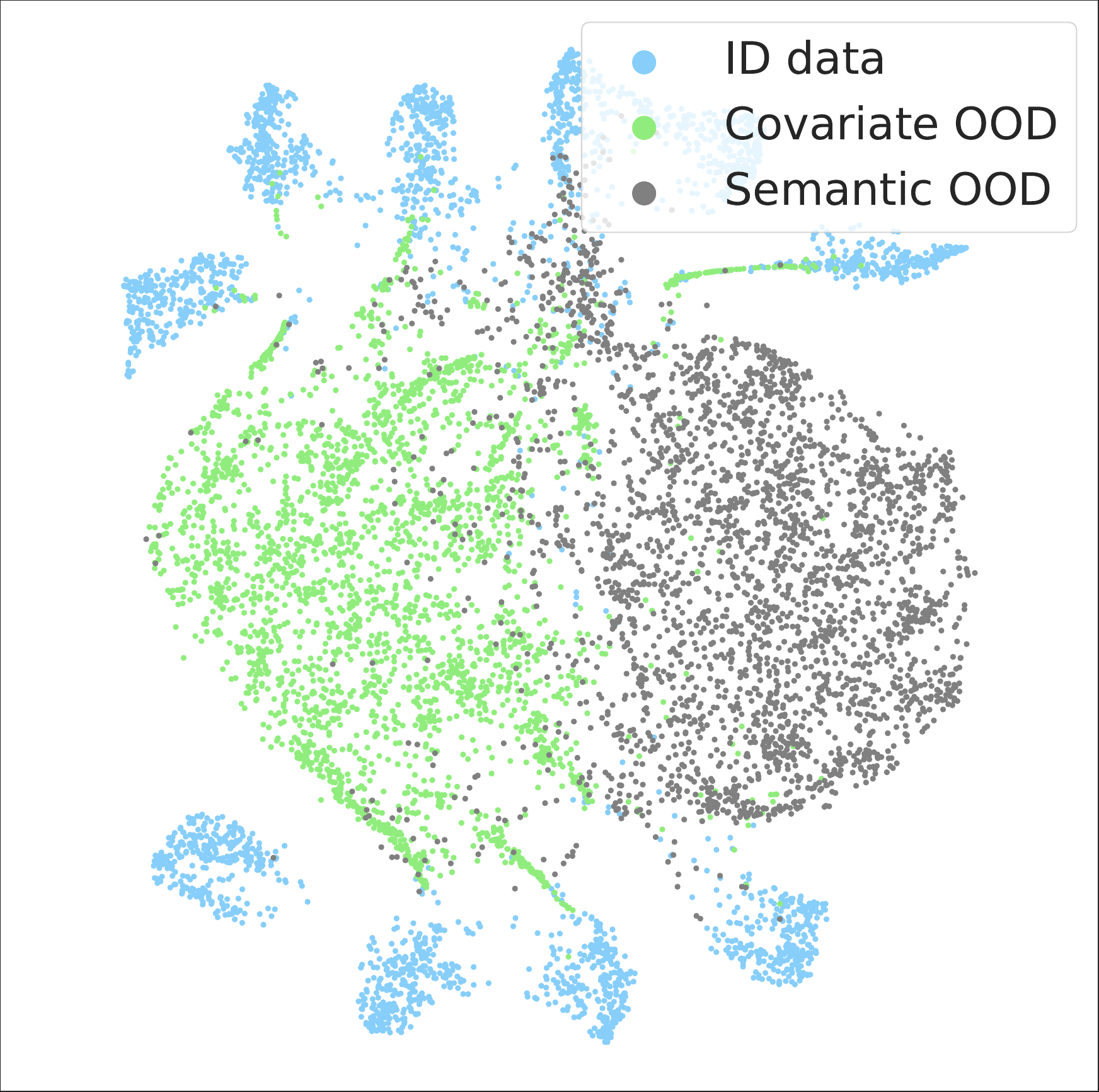}}
\hspace{1mm}
\subfigure[T-SNE of Ours]{\includegraphics[width=0.22\textwidth]{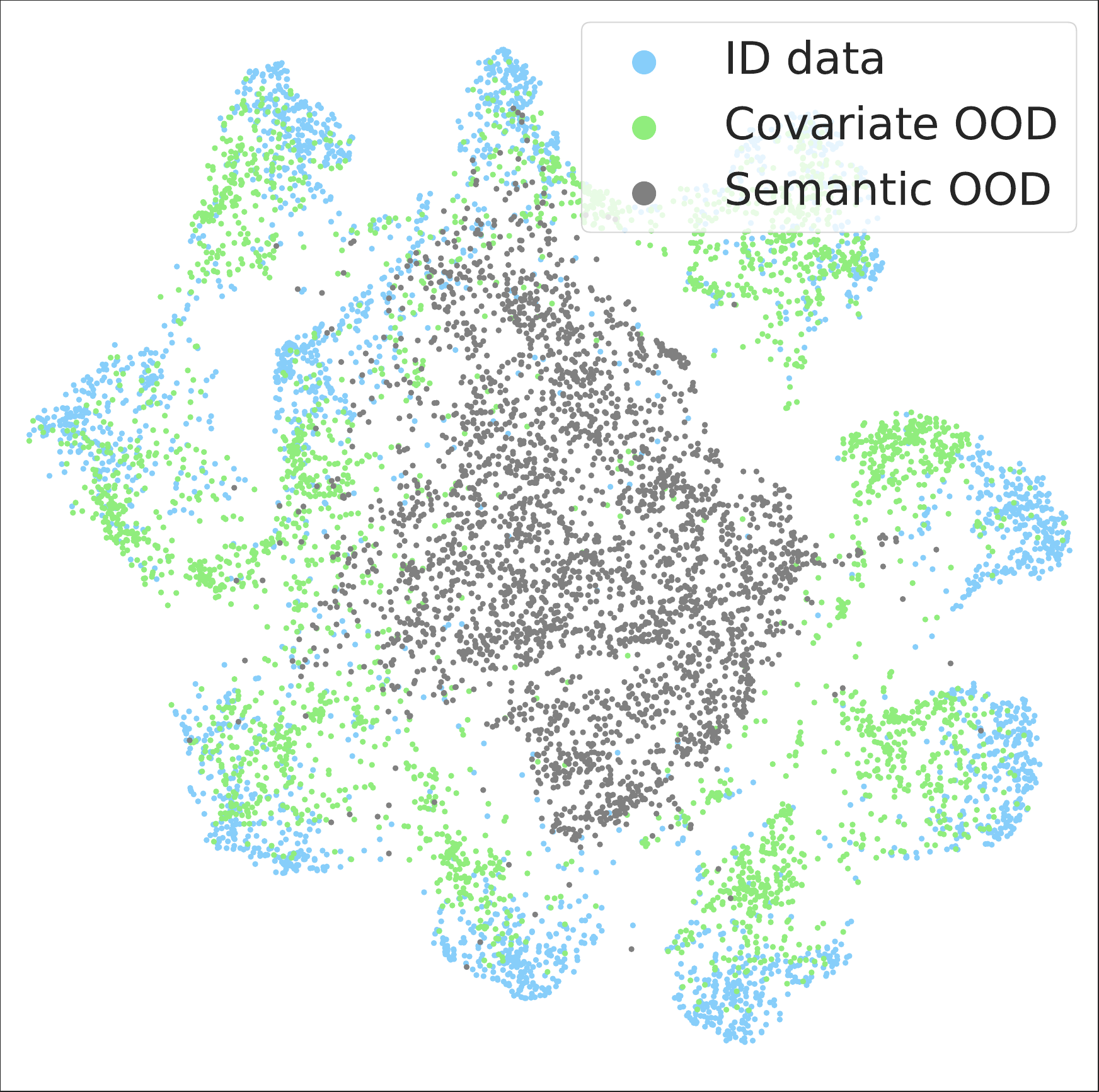}}
\vspace{-0.3cm}
\caption{
(a)-(b) Energy score distributions for WOODS vs. our method. Different colors represent the different types of test data: CIFAR-10 as $\Pin$ (\textcolor{blue}{blue}), CIFAR-10-C as $\mathbb{P}_\text{out}^\text{covariate}$ (\textcolor{green}{green}), and SVHN as $\mathbb{P}_\text{out}^\text{semantic}$ (\textcolor{gray}{gray}). (c)-(d): T-SNE visualization of the image embeddings using WOODS vs. our method.}
\vspace{-0.3cm}
\label{fig:tsne}
\end{figure*}

\begin{table}[t]
\centering
\caption{Ablations on mixing ratios $\pi_c$. We train on CIFAR-10 as ID, using CIFAR-10-C for $\mathbb{P}_\text{ood}^\text{covariate}$ and SVHN for $\mathbb{P}_\text{ood}^\text{semantic}$ (with fixed $\pi_s=0.1$). For our method, $\eta=-10$, which is chosen based on the validation procedure in Appendix~\ref{app:experimental_details}.}
\scalebox{0.85}{
\begin{tabular}{c|ccccc}
\toprule
\textbf{$\pi_c$} & \textbf{Method}   
& \textbf{OOD Acc.}$\uparrow$ & \textbf{ID Acc.}$\uparrow$ &\textbf{FPR}$\downarrow$ & \textbf{AUROC}$\uparrow$ \\
\midrule
{0.0} & \textsc{{Woods}} & {76.44} & {94.92} & {1.18} & {99.78} \\
{0.0} & \textsc{{Scone}} & {76.52} & {94.93} & {1.60} & {99.71} \\
\midrule
0.1 & \textsc{Woods} & 73.65 & 94.92 & 1.47 & 99.72 \\
0.1 & \textsc{Scone} & 78.27 & 94.88 & 10.61 & 97.91 \\
\midrule
0.2 & \textsc{Woods} & 65.55 & 95.02 & 2.19 & 99.57 \\
0.2 & \textsc{Scone} & 80.70 & 94.93 & 9.97 & 98.04 \\
\midrule
0.5 & \textsc{Woods} & 52.76 & 94.86 & 2.11 & 99.52 \\
0.5 & \textsc{Scone} & 84.69 & 94.65 & 10.86 & 97.84 \\
\midrule
0.9 & \textsc{Woods} & 52.84 & 94.81 & 1.80 & 99.57 \\
0.9 & \textsc{Scone} & 86.18 & 94.64 & 13.68 & 97.41 \\
\bottomrule
\end{tabular}%
}
\label{tab:ratio}%
\end{table}%

\paragraph{Effect of different mixing ratios.} In Table~\ref{tab:ratio}, we ablate the effect of $\pi_c$, which modulates the fraction of covariate OOD data in the mixture distribution $\Pwild$. For all settings, we contrast the performance without vs. with margin. The margin in each $\pi_c$ setting is validated using the strategy in Appendix~\ref{app:experimental_details}. Here we consistently use $\pi_s=0.1$, which  reflects the practical scenario that the majority of test data may remain in the known classes. 
{We primarily focus on evaluations where $\pi_c \ne 0$, since our problem setting uniquely introduces the covariate shift in the wild distribution. However, for completeness, we also include results when $\pi_c = 0$.}
We highlight a few interesting observations:~\textbf{(1)} without any enforced margin, the OOD generalization performance for WOODS~\cite{katz2022training} generally degrades with increasing $\pi_c$. This is likely due to the fact that a larger $\pi_c$  translates into more severe covariate shifts. For example, when $\pi_c=0.9$, the OOD classification accuracy decreases to 52.84\%. \textbf{(2)} Our method is overall more robust under large $\pi_c$ settings than the WOODS baseline. For instance, in a challenging case with $\pi_c=0.9$, \textsc{Scone} outperforms WOODS by \textbf{33.34\%}. Overall, these results demonstrate the benefits of \textsc{Scone} for both OOD generalization and detection.

\begin{table*}[t]
\centering
\caption{\small Results on ImageNet-100. We use ImageNet-100 as ID, ImageNet-100-C for $\mathbb{P}_\text{ood}^\text{covariate}$ and iNaturalist for $\mathbb{P}_\text{ood}^\text{semantic}$. }
\scalebox{0.8}{
\begin{tabular}{lcccc}
\toprule
\textbf{Method}  & \textbf{OOD Accuracy (ImageNet-100-C)} & \textbf{ID Accuracy (ImageNet-100)} & \textbf{FPR95} & \textbf{AUROC} \\
& (OOD generalization) & (ID generalization)& (OOD detection) & (OOD detection)  \\
& $\uparrow$ & $\uparrow$ & $\downarrow$ & $\uparrow$ \\
\midrule
WOODS~\cite{katz2022training} & 44.46 & 86.49 & 10.50 & 98.22 \\
SCONE (ours) & 65.34 & 87.64 & 27.13 & 95.66 \\
\bottomrule
\end{tabular}
}         
\label{tab:imagenet}
\end{table*}

\subsection{Qualitative Insights} \label{expe:visualize}

\textbf{Visualization of OOD score distributions.} We visualize the energy score distribution in Figure~\ref{fig:tsne} (a) and (b), for WOODS vs. our method, respectively. There are two salient observations: first, the energy scores for ID data indeed shift from -5 (without margin) towards more negative values (e.g., -16), suggesting the efficacy of our margin-based optimization. Moreover, the energy score distributions between $\mathbb{P}_\text{in}$ and $\mathbb{P}_\text{out}^\text{covariate}$ becomes more aligned than in WOODS. This can be attributed to the aligned feature representation, which we verify next.

\paragraph{Visualization of feature embeddings.} Figure~\ref{fig:tsne} shows t-SNE visualizations~\cite{van2008visualizing} of the penultimate-layer feature embedding, for WOODS (c) vs. our method (d). The embeddings are extracted from the test split. The \textcolor{blue}{blue} points denote the test ID set (CIFAR-10), the \textcolor{green}{green} points are the test samples from CIFAR-10-C, and the \textcolor{gray}{gray} points are from SVHN. This visualization suggests that embeddings of CIFAR and CIFAR-C become more aligned with our margin-based optimization, which arguably leads to improved OOD generalization performance. This observation corroborates our theoretical insight in Section~\ref{sec:theory}.

\subsection{Experiments on ImageNet}
\label{app:imagenet}
In this section, we provide additional large-scale results on the ImageNet benchmark. We use ImageNet-100 as the in-distribution data ($\mathbb{P}_\text{in}$), with labels provided in Appendix~\ref{app:datasets}. For the covariate-shifted OOD data ($\mathbb{P}_\text{out}^\text{covariate}$), we use ImageNet-100-C with Gaussian noise in the experiment. For the semantic-shifted OOD data, we use the high-resolution natural images from iNaturalist~\cite{van2018inaturalist}, with the same subset as MOS~\cite{huang2021mos}. The wild data $\mathbb{P}_\text{in}$ is a mixture of ID data, covariate-shifted data ($\pi_c = 0.5$), and semantic shifted data  ($\pi_s = 0.1$). 
{We fine-tune ResNet-34~\cite{he2016deep} (pre-trained on ImageNet) for 100 epochs, with an initial learning rate of 0.01 and a batch size of 64.}
Results in Table~\ref{tab:imagenet} suggest that our method can improve both ID and OOD accuracy compared to WOODS (the most competitive baseline).


\section{Related Works} \label{sec:rela}

\paragraph{Out-of-distribution detection} is of vital importance for machine learning models deployed in the open world. Recent advances in OOD detection can be broadly categorized into post hoc and regularization-based methods. In particular, post hoc methods~\cite{hendrycks2016baseline, liang2018enhancing, lee2018simple, liu2020energy, huang2021importance, sun2021react, sun2022out} focus on deriving test-time OOD scoring functions for a pre-trained classifier. Our proposed work is closer to another line of work~\cite{bevandic2018discriminative, hendrycks2018deep, malinin2018predictive, liu2020energy, du2021vos, ming2022posterior}, which addresses the OOD detection problem by training-time regularization. For example, models are encouraged to give predictions with lower confidence~\cite{hendrycks2018deep} or higher energies~\cite{liu2020energy}. These methods require access to a clean OOD dataset for training, which can be restrictive. To circumvent this, a recent work WOODS~\cite{katz2022training} first explored using wild mixture data consisting of the ID and semantically shifted OOD data $\mathbb{P}_\text{wild}:= (1-\pi) \mathbb{P}_\text{in} + \pi \mathbb{P}_\text{out}^\text{semantic}$. The crucial difference between our work and WOODS is whether the wild mixture data contains covariate-shifted data, which introduces new challenges not considered in prior work. As we show in this work, our formulation  {uniquely enables both OOD generalization and OOD detection}, in one coherent framework.

\paragraph{Out-of-distribution generalization} 
 is a fundamental problem in machine learning, which aims to generalize to covariate-shifted data without any sample from the target domain~\cite{muandet2013domain, arjovsky2019invariant, bahng2020learning,wang2022generalizing, xie2020n}. OOD generalization is more challenging compared to the classic domain adaptation problem~\cite{daume2006domain,blitzer2008learning,  ben2010theory,ganin2015unsupervised,tzeng2017adversarial,redko2019advances, kang2019contrastive,kumar2020understanding, wang2022embracing}, which assumes access to labeled samples from the target domain. 
 
To highlight a few works,
IRM~\cite{arjovsky2019invariant} and its variants~\cite{krueger2021out,ahuja2020invariant} aim to find invariant representation from different training environments via an invariant risk regularizer.
GroupDRO~\cite{sagawa2019distributionally} and Probabilistic Group DRO~\cite{ghosal2023pdgro} minimize the worst-case training loss over a set of groups.
\citet{zhou2020learning} propose generating outlier samples of a novel domain, which are then used for improving the generalization of the classifier. Besides algorithm innovation, benchmark efforts have also been pursued by DomainBed~\cite{gulrajani2020search} and 
OoD-Bench~\cite{ye2022ood}, which facilitates evaluation on OOD generalization. 
{OS-SDG~\cite{zhucrossmatch} relies on one source domain with labels to train the model, while our framework can exploit unlabeled wild data naturally arising in the wild, which is a mixed composition of three data distributions.}
Different from previous literature, we focus on improving OoD robustness in classifiers by learning from the wild mixture data and building an OOD detector at the same time. 
To the best of our knowledge, we are the first work that leverages wild data for both OOD generalization and OOD detection purposes. Our framework also uniquely allows leveraging covariate-shifted data freely arising in the wild, without requiring any labeling.

\paragraph{{Universal domain adaptation}} {aims to leverage labeled data from a related domain (source domain) and improve the model performance for the target domain, where there exists category gap for label sets between the source and target domains~\cite{you2019universal}. Several works have been proposed to address this problem~\cite{saito2020universal, fu2020learning, li2021domain, chen2022geometric, kundu2022subsidiary, chang2022unified, garg2022domain}. UAN~\cite{you2019universal} presents a universal adaptation network that exploits both the domain similarity and prediction uncertainty of each sample for promoting common-class adaptation. DANCE~\cite{saito2020universal} proposes a domain adaptative neighborhood clustering technique for category shift-agnostic adaptation via entropy optimization. The work in~\citet{chang2022unified} proposes a unified optimal transport-based framework to encourage both global cluster discrimination and local consistency of samples. 
Different from prior works, we leverage both labeled in-distribution data and unlabeled wild data when training our model. Such unlabeled wild data naturally arise in real-world environments and have not been considered in prior literature.}

\paragraph{Positive-Unlabeled (PU) learning} is a classic machine learning problem, which aims to learn classifiers from positive and unlabeled data~\cite{letouzey2000learning}. Multiple prior works have been proposed for discussing PU learning~\cite{hsieh2015pu, zhao2022dist, acharya2022positive, chapel2020partial, xu2021positive}. 
The work in~\cite{niu2016theoretical} proposes a theoretical comparison of positive-unlabeled learning against positive-negative learning based on the upper bounds of estimation errors.
~\citet{du2015convex} presents a convex formulation for PU learning by using different loss functions for positive and unlabeled samples.
Margin-based PU learning~\cite{gong2018margin} introduces a provable positive margin-based PU learning algorithm for classification under the truncated linear distributions.
There are two key differences between ours and PU learning: (1) PU learning only considers the task of distinguishing $\P_\text{out}$ (anomalous) and $\P_\text{in}$ (normal), not the task of doing classification simultaneously. We consider OOD detection which additionally requires learning a classifier for the distribution $\P_{\mathcal{X}\mathcal{Y}}$. (2) PU learning does not consider the generalization aspect under covariate-shifted OOD data, whereas our framework handles it in addition to semantic-shifted OOD data.


\section{Conclusion} \label{sec:conc}
In this study, we propose a novel framework \textsc{Scone} to jointly tackle the OOD generalization and OOD detection problems by leveraging wild data---a mixture of ID, covariate OOD, and semantic OOD data. Our framework offers practical advantages since the wild data is freely collectible in abundance, does not require any human annotation, and importantly, captures the environmental test-time OOD distributions under both covariate and semantic shifts.
We make use of such unlabeled wild data to train a binary OOD detector, and at the same time, enhance the generalization ability of the ID classifier. We provide new theoretical and empirical insights on the importance of enforcing a sufficient margin between the OOD decision boundary and ID data. Extensive experiments show that our framework can effectively improve both OOD generalization and detection performance.
We hope our framework will inspire both OOD generalization and OOD detection communities to tackle data shift problems synergistically.

\section*{Acknowledgement}
\label{sec:ack}

The work is supported in part by the AFOSR Young Investigator Award under No. FA9550-23-1-0184; Philanthropic Fund from SFF; Wisconsin Alumni Research Foundation (WARF); and faculty research awards/gifts from Google, Meta, and Amazon. Any opinions, findings, conclusions, or recommendations
expressed in this material are those of the authors and do not necessarily reflect the views, policies, or endorsements either expressed or implied, of the sponsors. The authors would also like to thank ICML
reviewers for their helpful suggestions and feedback.

\clearpage

\nocite{langley00}

\bibliography{main}
\bibliographystyle{icml2023}

\newpage
\appendix
\onecolumn

\section{Proof of Proposition \ref{prop:covcorrect}}
\label{app:proof}

In this {setting}, the energy function becomes
$E_\theta(\phi(\*x)) = -\log \bigl[e^{\frac12 \overline{f}_\theta(\phi(\*x))} + e^{-\frac12 \overline{f}_\theta(\phi(\*x))}\bigr]$. {If $E_\theta(\phi(\*x)) < \eta$, then} due to the symmetry of $e^u + e^{-u}$ {we have} $\abs{\overline{f}_\theta(\phi(\*x))} > -2\eta - 2\log 2$. Therefore, setting $\eta < 0$ effectively lower bounds the distance of ID points from the classification decision boundary in this setting.

Consider the value of $\overline{f}_\theta(\phi(\*x_c))$ for a covariate shifted point $\*x_c$ with label $y=1$. By assumption, there exists an ID point $\*x$, also with label $y=1$, satisfying $\norm{\phi(\*x_c) - \phi(\*x)}_2 < \delta$. We have
\begin{align*}
\overline{f}_\theta(\phi(\*x)) &= (\overline{f}_\theta(\phi(\*x)) - \overline{f}_\theta(\phi(\*x_c))) + \overline{f}_\theta(\phi(\*x_c)) \\
&\le \abs{\overline{f}_\theta(\phi(\*x)) - \overline{f}_\theta(\phi(\*x_c))} + \overline{f}_\theta(\phi(\*x_c)) \\
&\le L \norm{\phi(\*x) - \phi(\*x_c)}_2 + \overline{f}_\theta(\phi(\*x_c)) \\
&\le L \delta + \overline{f}_\theta(\phi(\*x_c)).
\end{align*}
Since $\alpha=0$, $\tau = 0$ and $y=1$, $\overline{f}(\phi(\*x)) = \abs{\overline{f}(\phi(\*x))} > -2\eta - 2\log2$, which combined with the above implies $\overline{f}(\phi(\*x_c)) > -2\eta - 2\log2 - L \delta$. A similar argument shows that if $y=2$, $\overline{f}(\phi(\*x_c)) < -(-2\eta - 2\log2 - L \delta)$. Therefore, one can set $\eta < - \log2 - \frac12 L \delta$ and ensure that $\*x_c$ is classified correctly, with $\overline{f}(\phi(\*x_c)) > 0$ when $y=1$ and $\overline{f}(\phi(\*x_c)) < 0$ when $y=2$. {We also immediately have an upper bound on $E_\theta(\*x_c)$, since 
\[E_\theta(\*x_c) \le -\frac12 \abs{\overline{f}_\theta(\*x_c)} \le \eta + \log2.\]
Under the setting $\eta < - \log2 - \frac12 L \delta$, $E_\theta(\*x_c) < - \frac12 L \delta < 0$, and so $g_\theta(\*x_c) = \textsc{in}$.
}

\section{Experimental Details}
\label{app:experimental_details}
\paragraph{Validation strategy for selecting $\eta$.}

Here we discuss how to choose the optimal margin parameter $\eta$ from $\{0, -0.1, -0.5, -1, -2, -10, -20, -50\}$. A major challenge is that one may not have access to a clean validation set of either $\mathbb{P}_\text{ood}^\text{covariate}$ or $\mathbb{P}_\text{ood}^\text{semantic}$. More realistically, one may have a separate unlabeled set sampled from the wild mixed distribution $\Pwild$. We thus leverage this \emph{mixed} dataset $\mathcal{D}_\text{val}$ for validation, and propose the following heuristic measurement that can help reliably select a good $\eta$:
$$ \tilde{\text{out}}\%= \frac{\sum_{\tilde{\*x}_i \in \mathcal{D}_\text{val}} \mathbbm{1}{\{g_\theta(\tilde{\*x}_i)=\text{out}\}}}{|\mathcal{D}_\text{val}|}.$$
This heuristic measures the fraction of samples in the validation set predicted as \textsc{out} by the OOD detector. We select $\eta$ based on a ``phase transition'' under this measurement. We exemplify this in Table~\ref{tab:validation}, based on our main experimental setting with CIFAR-10 as ID, CIFAR-10-C as Covariate-OOD, and SVHN as Semantic-OOD. The first phase (e.g., $\eta=0, 0.1, 0.5, 1, 2$) corresponds to an OOD detector that classifies ID on one side and remainder samples (including covariate shifted ones) to be on the other side. As the margin enlarges further (e.g., $\eta=-10$), the OOD detector primarily identifies $\mathbb{P}_\text{ood}^\text{semantic}$ as \textsc{out}, which matches more closely with the desired behavior of OOD detector. This behavior translates into a drop in $\tilde{\text{out}}\%$. We use the $\eta$ value corresponding to the drop as the selected margin parameter.

\begin{table}[h]
\centering
\vspace{-0.5cm}
\caption{Experimental results on CIFAR with different margin settings $\eta$. We train on CIFAR-10 as ID, using the same wild data with $\pi_c=0.5$ (CIFAR-10-C) and $\pi_s=0.1$ (SVHN).}
\scalebox{0.95}{
\begin{tabular}{c|ccccc}
\toprule
\textbf{margin}  
& \textbf{OOD Acc.}$\uparrow$ & \textbf{ID Acc.}$\uparrow$ &\textbf{FPR}$\downarrow$ & \textbf{AUROC}$\uparrow$ & $\tilde{\text{out}}$\%  \\
\midrule
No margin  & 52.77 & 94.87 & 2.11 & 99.52 & 58.49 \\
$\eta = -0.1$ & 53.24 & 94.87 & 2.16 & 99.52 & 58.32 \\
$\eta = -0.5$  & 54.22 & 94.85 & 2.31 & 99.49 & 58.16 \\
$\eta = -1$  & 55.55  & 94.88 & 2.56 & 99.45 & 57.53 \\
$\eta = -2$  & 58.47 & 95.00 & 3.19 & 99.35 & 55.54 \\
\rowcolor{mygray} $\eta = -10$ & 84.69 & 94.65 & 10.86 & 97.84 & 17.30 \\
$\eta = -20$ & 84.57 & 94.81 & 19.04 & 96.29 & 16.23 \\
$\eta = -50$ & 84.56 & 94.83 & 19.24 & 96.25 & 16.20 \\
\bottomrule
\end{tabular}%
}
\label{tab:validation}%
\end{table}%

\paragraph{Implementation details of baselines for OOD detection.} We use Wide ResNet~\cite{zagoruyko2016wide} with 40 layers and widen factor of 2. For evaluating the post hoc OOD detection baselines, we use the model trained with the CE loss on $\P_\text{in}$.  We employ the same pre-trained model from the github:~\url{https://github.com/wetliu/energy_ood}, which was trained on the complete CIFAR-10 dataset. 
Specifically, the model is trained using cross-entropy loss for 200 epochs. The learning rate is started at 0.1 and then decays by multiplier 0.1 at 100, 150, and 175 epochs. To facilitate easy comparison, the pre-trained model and baseline results are consistent with~\cite{katz2022training} (courtesy of Table 4).

\paragraph{Implementation details of baselines for OOD generalization.} 
For OOD generalization baselines, we use the same network architecture, Wide ResNet-40-2~\cite{zagoruyko2016wide}, and train from scratch using respective losses. The baselines in Table~\ref{tab:ood} are trained on CIFAR-10~\cite{krizhevsky2009learning} $50,000$ labeled training examples. We follow the default hyperparameter setting as in original papers whenever applicable.
For Mixup \cite{zhang2018mixup}, we follow the original paper and set hyperparameter $\alpha=1$ to control the strength of interpolation between feature-target pairs. The $\lambda$ for IRM~\cite{arjovsky2019invariant} baseline is set to 100 and the $\lambda$ for VREx~\cite{krueger2021out} is set to 10 for the penalty weights.
Following the original training configuration in WRN \cite{zagoruyko2016wide}, all the baselines are trained for 200 epochs with batch size 128. We use the SGD optimizer with an initial learning rate of 0.1. The learning rate decays by a factor of 10 after 60, 120, and 180 epochs. Weight decay is set to $10^{-4}$. All models are implemented in PyTorch 1.8.1. We evaluate the trained model on the CIFAR-10 test set (ID accuracy) and CIFAR-10-C (OOD accuracy).

\section{Details of Datasets} 
\label{app:datasets}

We provide a detailed description of the datasets used in this work below:

\textbf{MNIST} \cite{lecun1998mnist} is a large database of handwritten digits with 10 categories and is widely used in the field of machine learning. The MNIST contains $60,000$ training images and $10,000$ test images.

\textbf{CIFAR-10} \cite{krizhevsky2009learning} contains $60,000$ color images with 10 classes. The training set has $50,000$ images and the test set has
$10,000$ images.

\textbf{ImageNet-100} is composed by randomly sampled 100 categories from ImageNet-1K~\cite{deng2009imagenet}. This dataset contains the following classes: {\small n01498041, n01514859, n01582220, n01608432, n01616318, n01687978, n01776313, n01806567, n01833805, n01882714, n01910747, n01944390, n01985128, n02007558, n02071294, n02085620, n02114855, n02123045, n02128385, n02129165, n02129604, n02165456, n02190166, n02219486, n02226429, n02279972, n02317335, n02326432, n02342885, n02363005, n02391049, n02395406, n02403003, n02422699, n02442845, n02444819, n02480855, n02510455, n02640242, n02672831, n02687172, n02701002, n02730930, n02769748, n02782093, n02787622, n02793495, n02799071, n02802426, n02814860, n02840245, n02906734, n02948072, n02980441, n02999410, n03014705, n03028079, n03032252, n03125729, n03160309, n03179701, n03220513, n03249569, n03291819, n03384352, n03388043, n03450230, n03481172, n03594734, n03594945, n03627232, n03642806, n03649909, n03661043, n03676483, n03724870, n03733281, n03759954, n03761084, n03773504, n03804744, n03916031, n03938244, n04004767, n04026417, n04090263, n04133789, n04153751, n04296562, n04330267, n04371774, n04404412, n04465501, n04485082, n04507155, n04536866, n04579432, n04606251, n07714990, n07745940}.

\textbf{MNIST-C} \cite{mu2019mnist} is a corrupted version of MNIST data with different corruption types for benchmarking out-of-distribution robustness in computer vision.

\textbf{CIFAR-10-C} is algorithmically generated, following the previous leterature~\cite{hendrycks2018benchmarking}, from different corruptions for CIFAR-10 data including gaussian noise, defocus blur, glass blur, impulse noise, shot noise, snow, and zoom blur.

\textbf{ImageNet-100-C} is algorithmically generated with Gaussian noise based on \cite{hendrycks2018benchmarking} for the ImageNet-100 dataset~\cite{deng2009imagenet}.

\textbf{FashionMNIST} \cite{xiao2017fashion} consists of $70,000$ fashion products images from 10 categories, with $7,000$ images per category. There are $60,000$ training images and $10,000$ test images. The 10 categories include T-Shirt, Trouser, Pullover, Dress, Coat, Sandals, Shirt, Sneaker, Bad, and Ankle boots.

\textbf{SVHN} \cite{netzer2011reading} is a real-world image dataset obtained from house numbers in Google Street View images. This dataset $73,257$ samples for training, and $26,032$ samples for testing with 10 classes.

\textbf{Places365} \cite{zhou2017places} contains scene photographs and diverse types of environments encountered in the world. The scene semantic categories consist of three macro-classes: Indoor, Nature, and Urban.

\textbf{LSUN-C} \cite{yu2015lsun} and \textbf{LSUN-R} \cite{yu2015lsun} are large-scale image datasets that are annotated using deep learning with humans in the loop. LSUN-C is a cropped version of LSUN and LSUN-R is a resized version of the LSUN dataset, which has no overlap categories with the CIFAR dataset \cite{krizhevsky2009learning}.

\textbf{Textures} \cite{cimpoi2014describing} refers to the Describable Textures Dataset, which contains a large dataset of visual attributes including patterns and textures. The subset we used has no overlap categories with the CIFAR dataset \cite{krizhevsky2009learning}.

\textbf{iNaturalist} \cite{van2018inaturalist} is a challenging real-world dataset with iNaturalist species, captured in a wide variety of situations. It has 13 super-categories and 5,089 sub-categories. We use the subset from~\cite{huang2021mos} that contains 110 plant classes that no category overlaps with IMAGENET-1K~\cite{deng2009imagenet}.

\paragraph{Details of data split for OOD datasets.} For datasets with standard train-test split (e.g., SVHN), we use the original test split for evaluation. For other OOD datasets (e.g., LSUN-C), we use $70\%$ of the data for creating the wild mixture training data  as well as the mixture validation dataset. We use the remaining examples for test-time evaluation. For splitting training/validation, we use $30\%$ for validation and the remaining for training.

\begin{table*}[t]
\centering
\caption{Additional results. Comparison with competitive OOD detection and OOD generalization methods on CIFAR-10. For experiments using $\mathbb{P}_{\mathrm{wild}}$, we use $\pi_s = 0.5$, $\pi_c = 0.1$. For each semantic OOD dataset, we create corresponding wild mixture distribution $\mathbb{P}_\text{wild} := (1-\pi_s-\pi_c) \mathbb{P}_\text{in} + \pi_s \mathbb{P}_\text{out}^\text{semantic} + \pi_c \mathbb{P}_\text{out}^\text{covariate}$ for training. }
\scalebox{0.9}{
\begin{tabular}{lcccc|cccc}
\toprule
\multirow{2}[2]{*}{\textbf{Model}} & \multicolumn{4}{c}{{Texture $\mathbb{P}_\text{out}^\text{semantic}$, CIFAR-10-C $\mathbb{P}_\text{out}^\text{covariate}$}} & \multicolumn{4}{c}{{LSUN-R $\mathbb{P}_\text{out}^\text{semantic}$, CIFAR-10-C $\mathbb{P}_\text{out}^\text{covariate}$}}   \\
 & \textbf{OOD Acc.}$\uparrow$  & \textbf{ID Acc.}$\uparrow$ & \textbf{FPR}$\downarrow$ & \textbf{AUROC}$\uparrow$ & \textbf{OOD Acc.}$\uparrow$ & \textbf{ID Acc.}$\uparrow$ & \textbf{FPR}$\downarrow$ & \textbf{AUROC}$\uparrow$\\
\midrule
\emph{OOD detection}\\
\textbf{MSP} & 75.05 & 94.84 & 59.28 & 88.50 & 75.05 & 94.84 & 52.15 & 91.37 \\
\textbf{ODIN} & 75.05 & 94.84 & 49.12 & 84.97 & 75.05 & 94.84 & 26.62 & 94.57 \\
\textbf{Energy}  & 75.05 & 94.84 & 52.79 & 85.22 & 75.05 & 94.84 & 27.58 & 94.24 \\
\textbf{Mahalanobis}  & 75.05 & 94.84 & 15.00 & 97.33 & 75.05 & 94.84 & 42.62 & 93.23  \\
\textbf{ViM}  & 75.05 & 94.84 & 29.35 & 93.70 & 75.05 & 94.84 & 36.80 & 93.37 \\
\textbf{KNN} & 75.05 & 94.84 & 39.50 & 92.73 & 75.05 & 94.84 & 29.75 & 94.60 \\
\midrule
\emph{OOD generalization}\\
\textbf{ERM } & 75.05 & 94.84 & 52.79 & 85.22 & 75.05 & 94.84 & 27.58 & 94.24 \\
\textbf{Mixup } & 79.17 & 93.30 & 58.24 & 75.70 & 79.17 & 93.30 & 32.73 & 88.86 \\
\textbf{IRM } & 77.92 & 90.85 & 59.42 & 87.81 & 77.92 & 90.85 & 34.50 & 94.54 \\
\textbf{VREx }  & 76.90 & 91.35 & 65.45 & 85.46 & 76.90 & 91.35 & 44.20 & 92.55 \\
\midrule
\emph{Learning w. $\Pwild$}\\
\textbf{OE} &  44.71 & 92.84 & 29.36 & 93.93 & 46.89 & 94.07 & 0.7 & 99.78 \\
\textbf{Energy (w/ outlier)} & 49.34 & 94.68 & 16.42 & 96.46 & 32.91 & 93.01 & 0.27 & 99.94\\
\textbf{Woods} & 83.14 & 94.49 & 39.10 & 90.45 & 78.75 & 95.01 & 0.60 & 99.87 \\
\rowcolor{mygray} 
\textbf{Scone} (ours) & 85.56 & 93.97 & 37.15 & 90.91 & 80.31 & 94.97 & 0.87 & 99.79 \\
\bottomrule
\end{tabular}%
}
\label{tab:ood-part2}%
\end{table*}%

\section{ Results on Additional OOD Datasets}
\label{app:additional_ood}
In this section, we provide the main results on more OOD datasets including Textures \cite{cimpoi2014describing} and LSUN\_Resize \cite{yu2015lsun} in Table~\ref{tab:ood-part2}. We observe that our proposed approach achieves overall strong performance in OOD generalization and OOD detection on these additional OOD datasets. Particularly, we compare our method with post hoc OOD detection methods such as \texttt{MSP}~\cite{hendrycks2016baseline}, \texttt{ODIN}~\cite{liang2018enhancing}, 
\texttt{Energy}~\cite{liu2020energy}, 
\texttt{Mahalanobis}~\cite{lee2018simple}, 
\texttt{ViM}~\cite{wang2022vim},
and \texttt{KNN}~\cite{sun2022out}. These methods are all based on a model trained with cross-entropy loss, which suffers from limiting OOD generalization performance (75.05\%). In contrast, our method achieves an improved OOD generalization performance (e.g., 85.56\% when the wild data is a mixture of CIFAR-10, CIFAR-10-C, and Texture). 

We also compare our method with common OOD generalization baseline methods including \texttt{IRM}~\cite{arjovsky2019invariant}, \texttt{Mixup}~\cite{zhang2018mixup}, and \texttt{VREx}~\cite{krueger2021out}. Our method consistently achieves better results compared to these OOD generalization baselines. Lastly, we compare our method with strong OOD detection methods using $\mathbb{P}_\text{wild}$ such as \textsc{oe}~\cite{hendrycks2018deep}, energy-regularized learning~\cite{liu2020energy}, and WOODS~\cite{katz2022training}. Our method demonstrates strong performance on OOD generalization accuracy, which shows the effectiveness of our method for making use of the covariate OOD data.

\section{{Results on PACS}}
\label{app:pacs}

{In this section, we report results on the PACS dataset~\cite{li2017deeper} and present the comparisons with other baseline methods from the DomainBed. As shown in Table~\ref{tab:pacs},  we compare our method with a collection of common OOD generalization baselines, including \texttt{ERM}~\cite{vapnik1999overview}, \texttt{IRM}~\cite{arjovsky2019invariant}, 
\texttt{GroupDRO}~\cite{sagawa2019distributionally},
\texttt{I-Mixup}~\cite{zhang2018mixup}, 
\texttt{VREx}~\cite{krueger2021out}, 
\texttt{MLDG}~\cite{li2018learning}, 
\texttt{CORAL}~\cite{sun2016deep}, 
\texttt{MMD}~\cite{li2018domain}, 
\texttt{DANN}~\cite{ganin2016domain}, 
\texttt{CDANN}~\cite{li2018deep}, 
\texttt{MTL}~\cite{blanchard2021domain},
\texttt{SagNet}~\cite{nam2021reducing},
\texttt{ARM}~\cite{zhang2020adaptive}, 
\texttt{RSC}~\cite{huang2020self}.
Our approach SCONE ($86.4\%$) outperforms all of these OOD generalization baselines on the DomainBed benchmark.
}

\begin{table*}[t]
\centering
\scalebox{0.9}{\begin{tabular}{lccccc}
\toprule
\textbf{Algorithm}  & \textbf{Art painting} & \textbf{Cartoon} & \textbf{Photo} & \textbf{Sketch} & \textbf{Average Acc. (\%)}\\
\midrule
\textbf{ERM}~\cite{vapnik1999overview}  & 88.1 & 77.9 & 97.8 & 79.1 & 85.7  \\
\textbf{IRM}~\cite{arjovsky2019invariant}  & 85.0 & 77.6 & 96.7 & 78.5 & 84.4  \\
\textbf{GroupDRO}~\cite{sagawa2019distributionally}  & 86.4 & 79.9 & 98.0 & 72.1 & 84.1 \\
\textbf{I-Mixup}~\cite{wang2020heterogeneous,xu2020adversarial} & 86.5 & 76.6 & 97.7 & 76.5 & 84.3  \\
\textbf{VREx}~\cite{krueger2021out} & 86.0 & 79.1 & 96.9 & 77.7 & 84.9 \\
\textbf{MLDG}~\cite{li2018learning}  & 89.1 & 78.8 & 97.0 & 74.4 & 84.8  \\
\textbf{CORAL}~\cite{sun2016deep}  & 87.7 & 79.2 & 97.6 & 79.4 & 86.0 \\
\textbf{MMD}~\cite{li2018domain}   & 84.5 & 79.7 & 97.5 & 78.1 & 85.0  \\
\textbf{DANN}~\cite{ganin2016domain}  & 85.9 & 79.9 & 97.6 & 75.2 & 84.6   \\
\textbf{CDANN}~\cite{li2018deep}  & 84.0 & 78.5 & 97.0 & 71.8 & 82.8  \\
\textbf{MTL}~\cite{blanchard2021domain}  & 87.5 & 77.1 & 96.4 & 77.3 & 84.6 \\
\textbf{SagNet}~\cite{nam2021reducing} & 87.4 & 80.7 & 97.1 & 80.0 & 86.3  \\
\textbf{ARM}~\cite{zhang2020adaptive} & 86.8 & 76.8 & 97.4 & 79.3 & 85.1 \\
\textbf{RSC}~\cite{huang2020self} & 85.4 & 79.7 & 97.6 & 78.2 & 85.2 \\
\rowcolor{mygray}
\textbf{Ours} &  88.5 &  83.8 &  96.2 & 77.3  &  86.4 \\
\bottomrule
\end{tabular}}         
\vspace{-0.15cm}
\caption[]{{Comparison with OOD generalization algorithms on the PACS dataset from DomainBed benchmark. All methods are trained on ResNet-50. The model selection is based on a training domain validation set.}
}
\label{tab:pacs}
\end{table*}

{As shown in Table~\ref{tab:pacs-ood}, we summarize not only the OOD generalization performance but also the OOD detection performance on the PACS dataset. The results indicate that SCONE displays strong performance on both OOD generalization and detection tasks.}

\begin{table*}[ht]
\centering
\caption{{Results for both OOD generalization and detection tasks on the PACS dataset.}}
\scalebox{0.9}{
\begin{tabular}{lcccc}
\toprule
\textbf{Method}  & \textbf{OOD Accuracy} & \textbf{ID accuracy} & \textbf{FPR95} & \textbf{AUROC} \\
\midrule
Photo        & 96.23 & 99.68 & 2.57 & 99.38 \\
Art painting & 88.46 & 99.63 & 1.70 & 99.43 \\
Cartoon      & 83.75 & 99.52 & 0.63 & 99.68 \\
Sketch       & 77.25 & 99.01 & 17.80 & 96.34 \\
Average      & 86.42 & 99.46 & 5.68 & 98.71 \\
\bottomrule
\end{tabular}
}         
\label{tab:pacs-ood}
\end{table*}

\section{Results on Different Corruption Types}
\label{app:corruption}

In this section, we provide additional ablation studies of the different covariate shifts. {In Table~\ref{tab:corruption}, we evaluate our method under 19 different common corruptions such as \textit{gaussian noise}, \textit{defocus blur}, \textit{glass blur}, \textit{impulse noise}, \textit{shot noise}, \textit{snow}, \textit{zoom blur}, \textit{brightness}, etc.} 
We follow the default design and parameter setting as in the original paper~\cite{hendrycks2018benchmarking} for generating the corruptions. For our method with margin, $\eta$ is chosen based on the validation strategy in Appendix~\ref{app:experimental_details}. Our method is overall more robust under different covariate shifts than the WOODS baseline.

\begin{table*}[h]
\centering
\caption{Ablations on the different covariate shifts. We train on CIFAR-10 as ID, using CIFAR-10-C as $\mathbb{P}_\text{ood}^\text{covariate}$ and SVHN as $\mathbb{P}_\text{ood}^\text{semantic}$ (with  $\pi_c=0.5$ and $\pi_s=0.1$).}
\scalebox{0.95}{
\begin{tabular}{c|ccccc}
\toprule
\textbf{Covariate shift type} & \textbf{Method}   
& \textbf{OOD Acc.}$\uparrow$ & \textbf{ID Acc.}$\uparrow$ &\textbf{FPR}$\downarrow$ & \textbf{AUROC}$\uparrow$ \\
\midrule
Gaussian noise & \textsc{Woods} & 52.76 & 94.86 & 2.11 & 99.52 \\
Gaussian noise & \textsc{Scone} & 84.69 & 94.65 & 10.86 & 97.84 \\
\midrule
Defocus blur & \textsc{Woods} & 94.76 & 94.99 & 0.88 & 99.83 \\
Defocus blur & \textsc{Scone} & 94.86 & 94.92 & 11.19 & 97.81 \\
\midrule
Frosted glass blur & \textsc{Woods} & 38.22 & 94.90 & 1.63 & 99.71 \\
Frosted glass blur & \textsc{Scone} & 69.32 & 94.49 & 12.80 & 97.51 \\
\midrule
Impulse noise & \textsc{Woods} & 70.24 & 94.87 & 2.47 & 99.47 \\
Impulse noise & \textsc{Scone} & 87.97 & 94.82 & 9.70 & 97.98 \\
\midrule
Shot noise & \textsc{Woods} & 70.09 & 94.93 & 3.73 & 99.26 \\
Shot noise & \textsc{Scone} & 88.62 & 94.68 & 10.74 & 97.85 \\
\midrule
Snow & \textsc{Woods} & 88.10 & 95.00 & 2.42 & 99.54 \\
Snow & \textsc{Scone} & 90.85 & 94.83 & 13.22 & 97.32 \\
\midrule
Zoom blur & \textsc{Woods} &  69.15 & 94.86 & 0.38 & 99.91 \\
Zoom blur & \textsc{Scone} & 90.87 & 94.89 & 7.72 & 98.54 \\
\midrule
Brightness & \textsc{Woods} & 94.86 & 94.98 & 1.24 & 99.77 \\
Brightness & \textsc{Scone} & 94.93 & 94.97 & 1.41 & 99.74 \\
\midrule
Elastic transform & \textsc{Woods} & 87.89 & 95.04 & 0.37 & 99.92 \\
Elastic transform & \textsc{Scone} & 91.01 & 94.88 & 8.77 & 98.32 \\
\midrule
Contrast & \textsc{Woods} & 94.37 & 94.94 & 1.06 & 99.80 \\
Contrast & \textsc{Scone} & 94.40 & 94.98 & 1.30 & 99.77 \\
\midrule
Fog & \textsc{Woods} & 94.69 & 95.01 & 1.06 & 99.80 \\
Fog & \textsc{Scone} & 94.71 & 95.00 & 1.35 & 99.76 \\
\midrule
Frost & \textsc{Woods} & 87.25 & 94.97 & 2.35 & 99.55 \\
Frost & \textsc{Scone} & 91.94 & 94.85 & 10.08 & 98.03 \\
\midrule
Gaussian blur & \textsc{Woods} & 94.78 & 94.98 & 0.87 & 99.83 \\
Gaussian blur & \textsc{Scone} & 94.76 & 94.86 & 3.14 & 99.39 \\
\midrule
Jpeg & \textsc{Woods} & 84.35 & 94.96 & 1.73 & 99.68 \\
Jpeg & \textsc{Scone} & 87.87 & 94.90 & 8.14 & 98.49\\
\midrule
Motion blur & \textsc{Woods} & 82.54 & 94.79 & 0.47 & 99.88 \\
Motion blur & \textsc{Scone} & 91.95 & 94.90 & 9.15 & 98.18 \\
\midrule
Pixelate & \textsc{Woods} & 91.56 & 94.91 & 1.82 & 99.66 \\
Pixelate & \textsc{Scone} & 92.08 & 94.96 & 1.97 & 99.64 \\
\midrule
Saturate & \textsc{Woods} & 92.45 & 95.03 & 1.26 & 99.77 \\
Saturate & \textsc{Scone} & 93.38 & 94.92 & 10.27 & 97.88 \\
\midrule
Spatter & \textsc{Woods} & 92.38 & 94.98 & 1.94 & 99.64 \\
Spatter & \textsc{Scone} & 92.78 & 94.98 & 1.94 & 99.64 \\
\midrule
Speckle noise & \textsc{Woods} & 72.31 & 94.94 & 3.51 & 99.30 \\
Speckle noise & \textsc{Scone} & 88.51 & 94.83 & 11.05 & 97.82 \\
\bottomrule
\end{tabular}%
}
\label{tab:corruption}%
\end{table*}%


\end{document}